\documentclass[sigconf,screen]{acmart}

\usepackage{multirow}
\usepackage{booktabs}
\usepackage{threeparttable}
\usepackage{balance}
\usepackage{amsfonts}
\usepackage{algorithm}
\usepackage{algorithmic}
\usepackage{graphicx}
\usepackage{amsmath}
\usepackage{multirow}
\usepackage{makecell,rotating}
\usepackage{array}
\usepackage{booktabs}
\usepackage{threeparttable}
\usepackage{pifont}
\usepackage{fancyhdr}
\usepackage{lipsum}
\usepackage{setspace}
\usepackage[misc]{ifsym}

\newcommand{\INDSTATE}[1][1]{\STATE\hspace{#1\algorithmicindent}}

\usepackage{graphicx}
\usepackage{subfigure}
\usepackage{epstopdf}

\AtBeginDocument{%
  \providecommand\BibTeX{{%
    \normalfont B\kern-0.5em{\scshape i\kern-0.25em b}\kern-0.8em\TeX}}}

\copyrightyear{2022}
\acmYear{2022}
\setcopyright{acmcopyright}
\acmConference[MM '22] {Proceedings of the 30th ACM International Conference on Multimedia }{October 10--14, 2022}{Lisboa, Portugal.}
\acmBooktitle{Proceedings of the 30th ACM International Conference on Multimedia (MM '22), October 10--14, 2022, Lisboa, Portugal}
\acmPrice{15.00}
\acmISBN{978-1-4503-9203-7/22/10}
\acmDOI{10.1145/3503161.3547892}

\begin{document}
\title{MAPLE: Masked Pseudo-Labeling autoEncoder for Semi-supervised Point Cloud Action Recognition}




\author{Xiaodong Chen}
\authornote{This work is done when Xiaodong Chen is an intern at JD Explore Academy.}
\email{cxd1230@mail.ustc.edu.cn}
\affiliation{%
  \institution{University of Science and Technology of China}
}

\author{Wu Liu}
\authornote{Wu Liu is the corresponding author.} 
\email{liuwu1@jd.com}
\affiliation{%
  \institution{JD Explore Academy}
}

\author{Xinchen Liu}
\email{liuxinchen1@jd.com}
\affiliation{%
  \institution{JD Explore Academy}
}

\author{Yongdong Zhang}
\email{zyd73@ustc.edu.cn}
\affiliation{%
  \institution{University of Science and Technology of China}
}

\author{Jungong Han}
\email{jungong.han@aber.ac.uk}
\affiliation{%
  \institution{Aberystwyth University}
}

\author{Tao Mei}
\email{tmei@jd.com}
\affiliation{%
  \institution{JD Explore Academy}
}

\renewcommand{\shortauthors}{Xiaodong Chen et al.}

\begin{abstract}
Recognizing human actions from point cloud videos has attracted tremendous attention from both academia and industry due to its wide applications like automatic driving, robotics, and so on.
However, current methods for point cloud action recognition usually require a huge amount of data with manual annotations and a complex backbone network with high computation cost, which makes it impractical for real-world applications.
Therefore, this paper considers the task of semi-supervised point cloud action recognition.
We propose a Masked Pseudo-Labeling autoEncoder (\textbf{MAPLE}) framework to learn effective representations with much fewer annotations for point cloud action recognition.
In particular, we design a novel and efficient \textbf{De}coupled \textbf{s}patial-\textbf{t}emporal Trans\textbf{Former} (\textbf{DestFormer}) as the backbone of MAPLE.
In DestFormer, the spatial and temporal dimensions of the 4D point cloud videos are decoupled to achieve an efficient self-attention for learning both long-term and short-term features.
Moreover, to learn discriminative features from fewer annotations, we design a masked pseudo-labeling autoencoder structure to guide the DestFormer to reconstruct features of masked frames from the available frames.
More importantly, for unlabeled data, we exploit the pseudo-labels from the classification head as the supervision signal for the reconstruction of features from the masked frames.
Finally, comprehensive experiments demonstrate that MAPLE achieves superior results on three public benchmarks and outperforms the state-of-the-art method by 8.08\% accuracy on the MSR-Action3D dataset.
\footnote{See the project on \url{www.xiaodongchen.cn/MAPLE/}.}

\end{abstract}


\maketitle


\section{Introduction}
\label{sec:intro}

Point cloud videos, compared with 2D RGB videos, contain richer visual and geometric information for action recognition.
Researchers from academia and industry have recently focused on point cloud action recognition due to its wide potential applications in autonomous driving, industrial manufacturing, robotics, and so on~\cite{MSRAction:conf/cvpr/LiZL10,  JRDB:journals/tpami/martin2021jrdb, l:conf/aaai/LiuLGTM18, l:conf/cvpr/SunLBFMB22, l:liu2022recent, zheng2022gait3d}.
With the development of deep learning techniques such as deep neural networks and the transformer~\cite{Transformer:conf/nips/VaswaniSPUJGKP17}, significant progress has been made in this task~\cite{MeteorNet:conf/iccv/LiuYB19, pointnet:conf/cvpr/QiSMG17}.
However, the high computation cost and the requirement of large-scale annotated data hinder the practical application of point cloud action recognition.

\begin{figure}[t]
  \centering
    \includegraphics[width=0.99\linewidth]{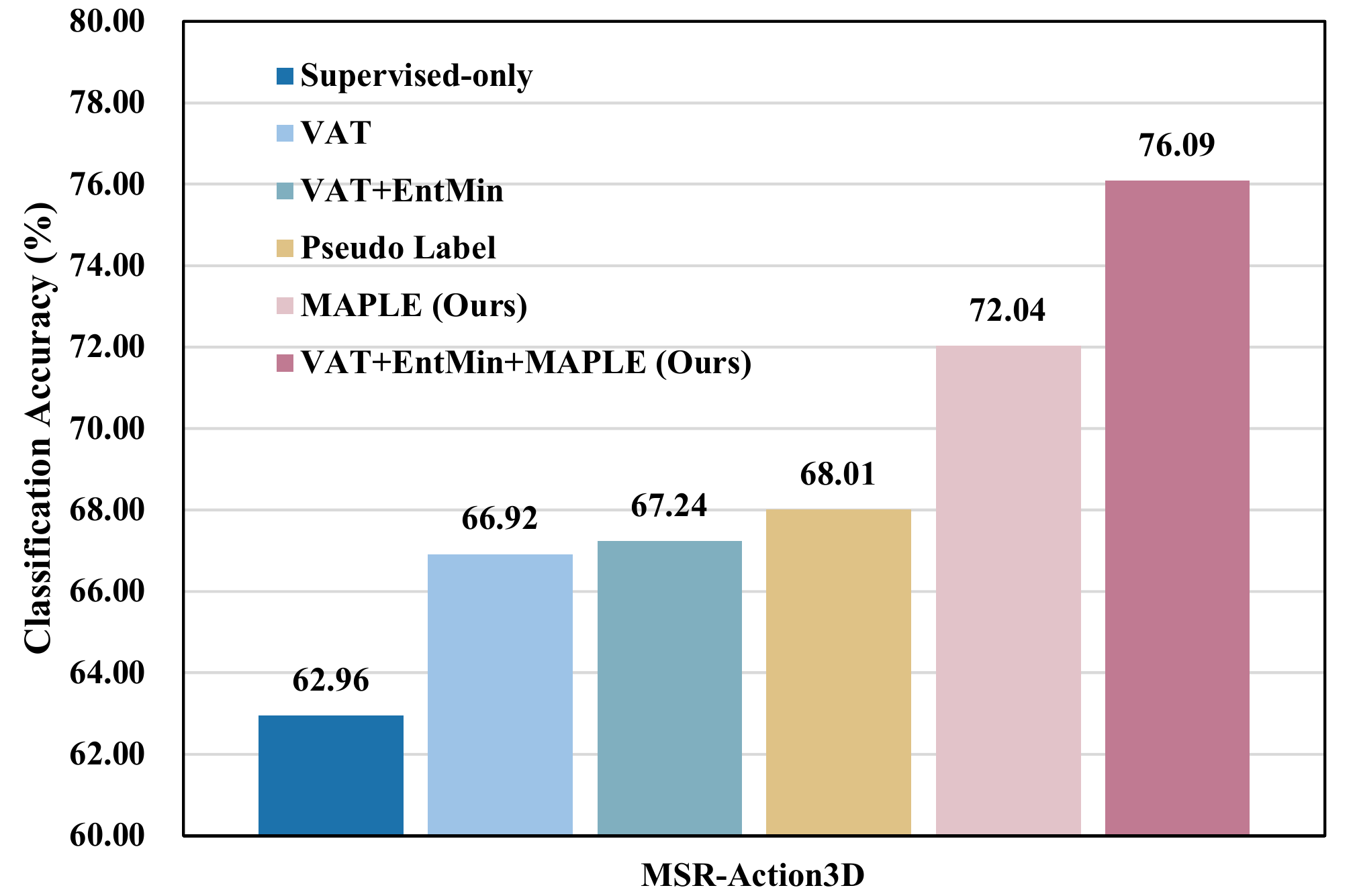}
    \caption{Comparisons of different semi-supervised methods, i.e., VAT~\cite{VAT:journals/pami/MiyatoMKI19}, EntMin~\cite{EntMin:conf/nips/GrandvaletB04}, and Pseudo Label~\cite{Pseudo:conf/icml/lee2013pseudo} on the MSR-Action3D dataset in terms of classification accuracy.}
    \label{fig:teaser1}
\vspace{-5mm}
\end{figure}

To recognize human actions in point cloud videos, the mainstream methods are divided into three categories. The first one~\cite{3DV:conf/cvpr/Wang0XJCZY20} is to convert the point cloud video into a series of ordered voxels and then apply traditional grid-based convolutions to these voxels.
The second type of method ~\cite{MeteorNet:conf/iccv/LiuYB19, pointnet2:conf/nips/QiYSG17} is to model and track local points with pointnet-based~\cite{pointnet:conf/cvpr/QiSMG17} models such as MeteorNet~\cite{MeteorNet:conf/iccv/LiuYB19}.
However, these two types of methods suffer from low computational efficiency and point-tracking errors~\cite{MeteorNet:conf/iccv/LiuYB19} respectively.
To address these problems, He et al.~\cite{P4D:conf/cvpr/Fan0K21} proposed the third method that extracts short-term local features by 4D convolutions and models long-term global information with the transformer. 
Nevertheless, the transformer-based methods usually require large-scale labeled data for training as the transformer has a larger model capacity~\cite{captity:journals/corr/abs-2108-13002}.
Although a large amount of point cloud videos can be easily obtained, labeling point cloud often needs much more cost on manual annotations compared to 2D RGB videos, which hinders the application of these methods.
Therefore, this paper focuses on the task of semi-supervised point cloud action recognition, which aims to reduce the reliance on manual labels in point cloud action recognition using a more efficient model.

Although the accuracy of current methods has been greatly improved, designing an annotation and computation-efficient framework for point cloud action recognition still faces several challenges.
First of all, due to the noises and ambiguity of point clouds, how to learn discriminative features and model the spatial-temporal patterns from the point clouds is a great challenge.
In image classification, researchers have studied combining techniques of CNNs with the transformer to improve the capability while reducing the computation complexity of the transformer models.
For example, Swin-Transformer~\cite{Swin:conf/iccv/LiuL00W0LG21} greatly enhances the capacity of the transformer while improving its efficiency by the shifted windows, which demonstrates the potential of the transformer in the modality of RGB images. 
However, such models are limited on the point cloud action recognition task due to the irregularity of the point clouds.

The other challenge is how to reduce the dependence on manual annotations while preserving the capability of the learned feature representations through an appropriate learning paradigm.
A common and effective learning framework is semi-supervised learning, which has rich applications in the field of image recognition and video understanding. 
Besides, Self-Supervised Learning (SSL) is also a powerful learning framework that exploits the generalizable representations from unlabeled data.
In particular, some recent research on the field of SSL~\cite{MAE:journals/corr/abs-2111-06377, MoCo:conf/cvpr/He0WXG20, MoCo2:journals/corr/abs-2003-04297} has shown excellent results, yet self-supervision alone is still insufficient due to its limited practical applicability.
To solve this dilemma, Zhai et al.~\cite{S4L:conf/iccv/BeyerZOK19} propose a new learning framework that combines self-supervised learning and semi-supervised learning and becomes a new paradigm in the semi-supervised field. 
However, limited by the invariance and unordered properties~\cite{pointnet:conf/cvpr/QiSMG17} of the point clouds, such methods cannot be directly applied to point cloud action recognition.

To overcome these challenges, we propose a novel learning framework named Masked Pseudo-Labeling autoEncoder (\textbf{MAPLE}) for point cloud action recognition. 
It introduces an autoencoder into the semi-supervised point cloud action recognition task. 
We also design an efficient transformer-based model named \textbf{De}coupled \textbf{s}patial-\textbf{t}emporal Trans\textbf{Former} (\textbf{DestFormer}) for this new learning framework. 
Based on this DestFormer backbone, we design an encoder-decoder structure for MAPLE.
It consists of a spatial extractor for learning short-term global features of actions, a temporal encoder for learning long-term action information, and a temporal decoder for feature reconstruction. 
To learn action information from the unlabeled action sequences, we reconstruct the masked action sequence with a highly masking ratio (e.g. 75 \%) during the training process.
However, directly reconstructing the video action sequence tends to result in the non-convergence of the model training and a decrease in classification performance.
Inspired by the knowledge distillation~\cite{Distillation:journals/corr/HintonVD15}, we  implicitly reconstruct the masked input sequence with the pseudo-label generated by the classification head to avoid this situation.
Besides, to exploit the potential of MAPLE, we further combine MAPLE with the classical semi-supervised learning methods to improve the performance of semi-supervised point cloud action recognition. 

We conduct extensive experiments with our MAPLE framework for semi-supervised point cloud action recognition on three widely-used datasets: MSR-Action3D (MSR3D)~\cite{MSRAction:conf/cvpr/LiZL10}, NTU RGB+D 60 (NTU60)~\cite{NTU60:conf/cvpr/ShahroudyLNW16}, and NTU RGB+D 120 (NTU120)~\cite{NTU120:journals/pami/LiuSPWDK20}. 
As shown in Fig.~\ref{fig:teaser1}, we make remarkable progress in the mainstream datasets compared to previous methods, e.g., VAT~\cite{VAT:journals/pami/MiyatoMKI19}, EntMin~\cite{EntMin:conf/nips/GrandvaletB04}, and Pseudo Label~\cite{Pseudo:conf/icml/lee2013pseudo}).
The MAPLE framework achieves the new state-of-the-art performance of the semi-supervised point cloud action recognition.

In summary, the contributions of this paper are three-fold:
\begin{itemize}
\item We present one of the first attempts toward semi-supervised point cloud action recognition which aims to learn efficient action representations from massive point cloud videos with fewer manual annotations. 
\item We design a \textbf{De}coupled \textbf{s}patial-\textbf{t}emporal Trans\textbf{Former}, named \textbf{DestFormer}, as the backbone of our semi-supervised learning framework, which decouples the spatial and temporal dimensions of the 4D point cloud videos for achieving a more efficient and effective self-attention.
\item We propose a Masked Pseudo-Labeling autoEncoder (\textbf{MAPLE}) framework for learning a generalizable and discriminative classifier through reconstructing motion features of masked frames from the available action frames.
\end{itemize}

\begin{figure*}[t]
  \centering
  \vspace{0mm}
    \includegraphics[width=0.99\linewidth]{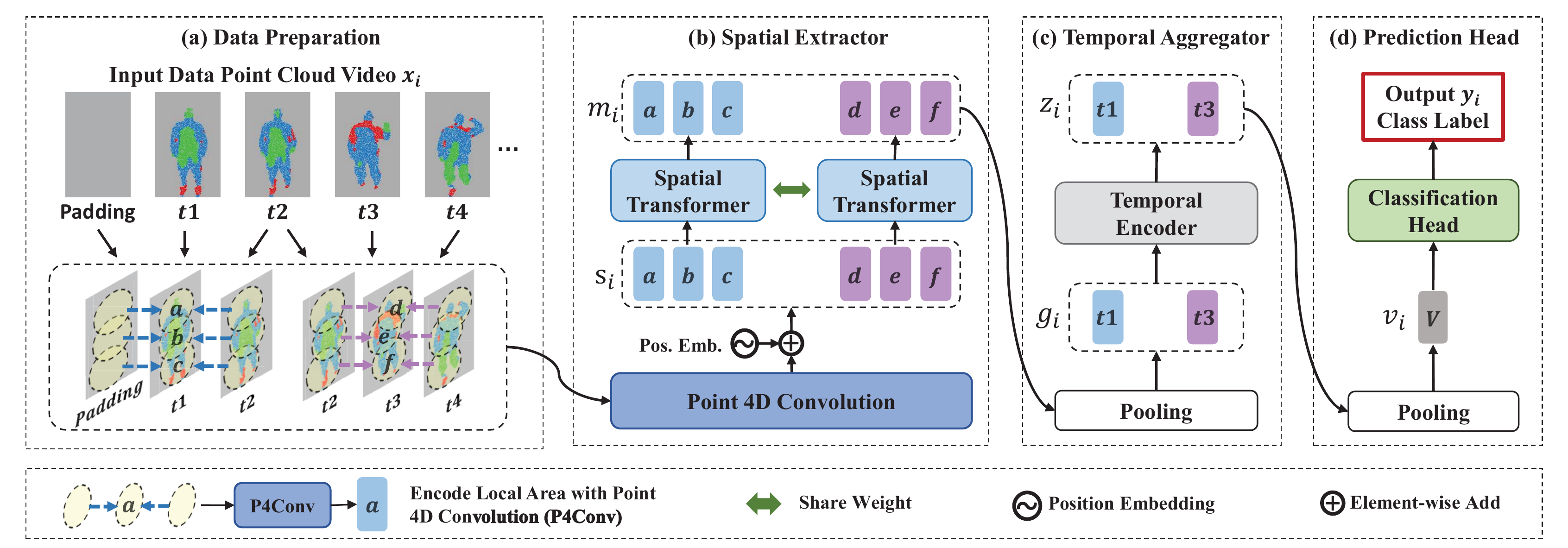}
\caption{The detail of DestFormer. (a) \textit{Data Preparation}: we construct some local areas (e.g. ``a'') on adjacent frames (e.g. ``t1'', ``t2'') from the input $x_i$ as what P4Conv~\cite{P4D:conf/cvpr/Fan0K21} do. (b) \textit{Spatial Extractor}: we adopt P4Conv for modeling short-time local information and feed the output $s_i$ frame by frame into a spatial transformer for extracting the merged local feature $m_i$. (c) \textit{Temporal Aggregator}: we generate the short-term global feature $g_i$ through the pooling layer and aggregate the long-term global information with the temporal encoder. (d) \textit{Prediction Head}: we project the global feature $v_i$ into label space via the classification head. }
  \label{fig:teaser2}
\vspace{-3mm}
\end{figure*}

\section{Related Work}
\label{sec:relatedwork}
\textbf{Point cloud action recognition} is a popular topic of video understanding in computer vision, which aims to help the machines understand the 3D world. 
Based on the characteristics of previous methods, three main categories have been distinguished. 
The first one is mainly based on voxels obtained from point clouds. 
3D dynamic voxel (3DV)~\cite{3DV:conf/cvpr/Wang0XJCZY20} brought the voxelization of point clouds into point cloud action recognition, via temporal rank pooling. 
It learned both action information through voxels and appearance through point cloud to encode the temporal information.
The second type of method is directly performed on the original point cloud with pointnet-based~\cite{pointnet:conf/cvpr/QiSMG17} models. 
For example, pointnet++~\cite{pointnet2:conf/nips/QiYSG17} borrowed the idea of local receptive fields to extract the spatial information of point clouds. MeteorNet~\cite{MeteorNet:conf/iccv/LiuYB19} further constructed the concept of spatial-temporal neighborhoods based on pointnet++ and determined the neighborhoods with direct grouping or chained-flow grouping.
The last category adopts the data-hungry transformer-based model in point cloud action recognition. 
P4Transformer~\cite{P4D:conf/cvpr/Fan0K21} directly modeled the action and appearance information of the whole video while effectively discarding the requirement of point-tracking used in MeteorNet and the complex calculations of voxelization. 
Similarly, $PST^2$~\cite{PST2:conf/wacv/WeiLXKG22} captured the spatial-temporal context information with the Spatial-temporal self-attention module. 
In this paper, our DestFormer belongs to the last category, but has 
less Floating-point Operations (FLOPs), powerful model capability, and less annotation dependence.

\textbf{Semi-Supervised Learning} is an important research topic in the field of pattern recognition and machine learning, which learns knowledge from the datasets including the much more set of unlabeled data and fewer labeled data. 
The theory and algorithms of semi-supervised learning were first summarized by Chapelle~\cite{SSL:books/mit/06/CSZ2006} in 2006 and Zhu~\cite{SSL:books/mit/08/Zhu2008} in 2008. 
The semi-supervised learning methods can be divided into two categories: the inductive methods and the transductive methods. 
The inductive method~\cite{SSL2010:conf/icml/LiuHC10, SSL2015:journals/kais/TrigueroGH15, SSL2017:journals/pr/SheikhpourSGC17} usually constructs a classifier for predicting the label of the whole dataset, including both the labeled and unlabeled data. 
By way of illustration, Grandvalet and Bengio~\cite{EntMin:conf/nips/GrandvaletB04} optimized the pseudo label generated by unlabeled data with conditional entropy minimization. 
Miyato et al.~\cite{VAT:journals/pami/MiyatoMKI19} added the small perturbations to the original input and constrained the output of unlabeled data with regularization. 
Another transductive method~\cite{SSL2009:conf/icml/JebaraWC09, SSL2012:journals/pieee/LiuWC12, SSL2014:series/synthesis/2014Subramanya} was always performed on the graph-based model. 
Different from the inductive methods, the transductive methods never produce a classifier for prediction. 
It usually defines a graph for all input data and encodes the relationship between the pairwise data points.

\textbf{Self-Supervised Learning} completely abandons the reliance on manual labels by adopting the input itself as supervision, thus making great progress on representation learning in the last few years. 
Through Liu's research~\cite{ssl2020:journals/corr/abs-2006-08218} on SSL, its main methods can be divided into three categories: generative SSL, contrastive SSL, and generative-contrastive SSL.
The generative SSL trains a generator consisting of an encoder and decoder to reconstruct the input data. 
Its represent research in natural language processing is GPT~\cite{GPT:journals/corr/radford2018improving} and BERT~\cite{BERT:conf/naacl/DevlinCLT19}, which predict the discarded content with the partially abandoned input sequence. 
In the field of computational vision, especially in the area of image classification and image generation, PixelCNN~\cite{PixelCNN:conf/icml/OordKK16}, VQ-VAE-2~\cite{vqvae:conf/nips/RazaviOV19}, and MAE~\cite{MAE:journals/corr/abs-2111-06377} successfully used the whole input image as the self-supervised target.
Compared to generative SSL, the motivation of contrastive SSL is to measure the similarity of different inputs (e.g., mutual information maximization and instance discrimination). 
Its influential work includes MoCo~\cite{MoCo:conf/cvpr/He0WXG20, MoCo2:journals/corr/abs-2003-04297}, BYOL~\cite{BYOL:conf/nips/GrillSATRBDPGAP20}, and SimCLR~\cite{SimCLR:conf/icml/ChenK0H20}.
As for the generative-contrastive SSL, most works focus on learning knowledge from unlabeled data with generative adversarial networks.

This paper links the inductive semi-supervised algorithm (e.g. Pseudo Label~\cite{Pseudo:conf/icml/lee2013pseudo}) with the generative self-supervision (e.g. MAE~\cite{MAE:journals/corr/abs-2111-06377}). 
Our MAPLE uses the short-term pseudo label instead of the short-term features as the reconstruction target.
By this means, it can achieve the stable reconstruction of masked frames and learn generalizable features from unlabeled point cloud videos.

\section{The Proposed MAPLE Framework}
\label{sec:method}

\begin{figure*}[t]
  \centering
  \vspace{0mm}
    \includegraphics[width=0.99\linewidth]{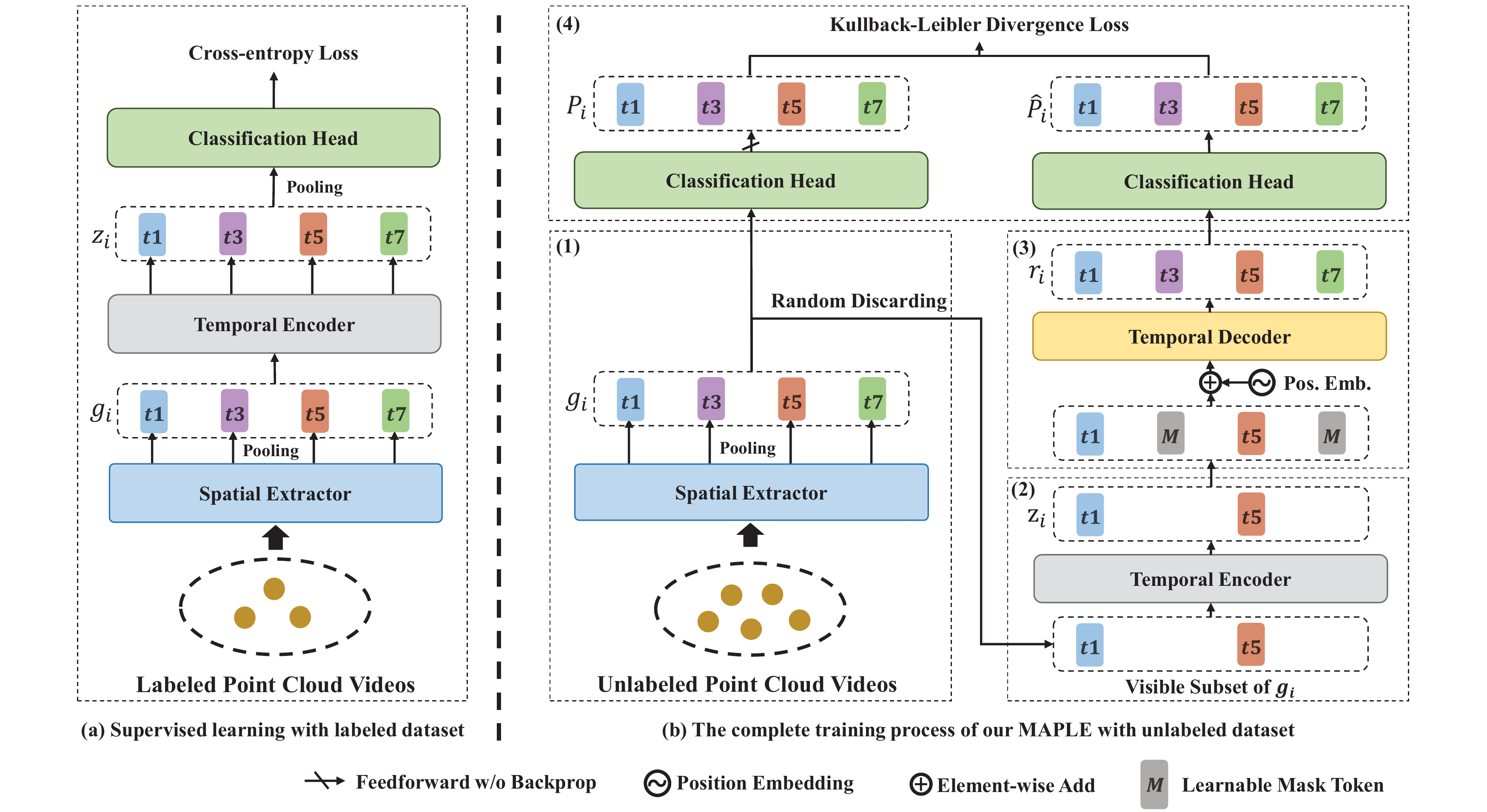}
    \caption{The detail of our MAPLE. 
(\textit{a}) Adopting our DestFormer backbone and the cross-entropy loss for the supervised training. 
(\textit{b}) The complete training process of MAPLE: (1) The spatial extractor encodes the input video as the short-term global feature $g_i$. (2) After randomly discarding the short-term global feature $g_i$, the temporal encoder projects the visible subset of $g_i$ as the latent representation $z_i$. (3) The temporal decoder is responsible for reconstructing $r_i$ from the latent representation $z_i$ and the mask tokens $M$. (4) The classification head generates the pseudo-label $P_i$ and $\hat{P}_i$ as our reconstruction target. Note that the modules here with the same colors share weights.}
  \label{fig:teaser3}
\vspace{-3mm}
\end{figure*}

In this section, we declare the detailed framework of our MAPLE. Our MAPLE consists of a Decoupled spatial-temporal TransFormer (DestFormer) backbone, as shown in Fig.~\ref{fig:teaser2}. The DestFormer takes the point cloud videos $x_i$ as input and adopts the spatial extractor, temporal aggregator and prediction head back-to-back for extracting the global feature $v_i$ and predicting the final class label $y_i$. On this basis, our MAPLE builds a masked autoencoder learning framework for semi-supervised action recognition, as shown in Fig.~\ref{fig:teaser3}. It consists of an encoder-decoder structure and implicitly reconstructs the masked input feature with the pseudo-label generated by the classification head. Before describing our MAPLE in detail, we first declare the necessary notations and definitions of the semi-supervised point cloud action recognition task.

\subsection{Preliminary}
\label{subsec:preliminary}
The task of point cloud action recognition is, given a point cloud video of humans, to predict the human behavior and actions in the video. Semi-supervised point cloud action recognition is consistent with the point cloud action recognition task in the inference phase, but usually adopts a different paradigm in the training phase as follows. 

1) We have a dataset $D$, which contains a labeled subset $D_l=\{(x_i,y_i)\}$ and an unlabeled subset $D_u=\{x_j\}$. Both $D_l$ and $D_u$ are sampled i.i.d. from the same distribution $p(x)$ and in general the size of $D_l$ is smaller than the size of $D_u$. 
Let $x_i = \{X_t \in \mathbb{R}^{(3+C)\times N}\}_{t=1}^{T}$ denote the matrix sequence of a point cloud video, where N indicates the number of points in each frame, $T$ indicates the number of frames in this video, $\mathbb{R}^{3}$ and $\mathbb{R}^{C}$ indicates the spatial coordinates and features dimension of one point. 
It is worth noting that there are no point cloud features (ie., $C=0$) in the given dataset (NTU RGB+D 60~\cite{NTU60:conf/cvpr/ShahroudyLNW16}, NTU RGB+D 120~\cite{NTU120:journals/pami/LiuSPWDK20} and MSR-Action3D~\cite{MSRAction:conf/cvpr/LiZL10}).

2) The training of model $f_\theta(\cdot)$ has an optimization function of the following form:
\begin{equation}
    \min_{\theta}  L_l(D_l,\theta) + \alpha L_u(D_u,\theta),
\label{equ:equ1}
\end{equation}
where $L_l$ is the loss function (e.g., cross-entropy loss, mean squared error loss, Hinge loss, etc.) for classification of the labeled dataset $D_l$, and $L_u$ is the optimization objective designed for unlabeled dataset $D_u$ (the design of this function varies from paper to paper, and we discuss our designed $L_u$ in later subsection.), $\alpha$ is the positive scalar weight for $L_u$ and $\theta$ is the learnable parameters of $f_\theta(\cdot)$.

\subsection{Decoupled Spatial-temporal TransFormer}
\label{subsec:P4ST-Trans}

This subsection presents our backbone for point cloud action recognition named Decoupled spatial-temporal TransFormer (DestFormer).
The design of DestFormer is based on P4Transformer~ \cite{P4D:conf/cvpr/Fan0K21}, and its main purpose is to serve as a basic backbone for the semi-supervised learning framework MAPLE and learn discriminative motion representations with fewer annotations. As shown in Fig.~\ref{fig:teaser2}, the DestFormer consists of four parts: data preparation, spatial extractor, temporal aggregator, and prediction head. Note that embedding features in figures with different colors (e.g. ``t1'' and ``t3'') correspond to the different keyframes of the input action sequence.

\textbf{Data Preparation.} For the input point cloud video $x_i$, we construct some local areas (e.g. ``a'',``b'',``c'') on adjacent frames (e.g. ``t1'',``t2'') as what Point 4D Convolution (P4Conv)~\cite{P4D:conf/cvpr/Fan0K21} do. The calculation of the local areas is based on the Farthest Point Sampling (FPS) algorithm~\cite{pointnet2:conf/nips/QiYSG17}, and the exhaustive calculation process is declared in~\cite{P4D:conf/cvpr/Fan0K21}.

\textbf{Spatial Extractor (SE)}. SE is designed to extract the short-term global feature $g_i=\{G_{\zeta\cdot(t-1)+1} \in \mathbb{R}^{D}\}_{t=1}^{T/\zeta}$ from the local areas.
We first extracts the short-term local feature $s_i = \{S_{\zeta\cdot(t-1)+1} \in \mathbb{R}^{D\times (N/\kappa)}\}_{t=1}^{T/\zeta}$ through P4Conv~\cite{P4D:conf/cvpr/Fan0K21}, where $D$ is the dimension of short-term local feature, $\kappa \geq 1.0$ denotes the spatial scaling rate and $\zeta \geq 1.0$ denotes the spatial scaling rate. P4Conv plays the role of aggregating the local information between adjacent $\zeta$ frames.
After that, we feed the short-term local feature $s_i$ frame by frame into the Spatial Transformer (ST) modules for extracting the merged short-term local feature  $m_i = \{M_{\zeta\cdot(t-1)+1} \in \mathbb{R}^{D\times (N/\kappa)}\}_{t=1}^{T/\zeta}$ which aggregate the information of different spatial part.

\textbf{Temporal Aggregator (TA)}. TA consists of a pooling layer and a transformer-based ~\cite{Transformer:conf/nips/VaswaniSPUJGKP17} Temporal Encoder (TE). we first prepare the short-term global feature $g_i$ from the merged short-term local feature  $m_i$ through the pooling layer (e.g. maximum pooling). Then we aggregate the long-term global $z_i$ from the short-term global feature $g_i$ with our TE module.

\textbf{Prediction Head}. Following the TA module is the pooling layer (e.g. maximum pooling) and classification head, which consists of Layer Normalization layers (LayerNorm), linear layers, and Gaussian Error Linear Units (GELUs). Its role is to project the global feature $v_i$ into the label space and generate the corresponding pseudo labels for classification.


\subsection{Masked Pseudo-labeling Autoencoder}
\label{subsec:MAPLE}
This subsection elaborates our Masked Pseudo-Labeling autoEncoder (MAPLE) framework for semi-supervised point cloud action recognition.
As shown in the left part of Fig.~\ref{fig:teaser3}, we adopt our DestFormer backbone and the cross-entropy loss for the supervised training with labeled point cloud videos. 
The right part of Fig.~\ref{fig:teaser3} shows the complete training process of our MAPLE with unlabeled point cloud videos.
Similar to the reconstruction process of autoencoders, the spatial extractor encodes the point cloud videos as the short-term local embedding feature $g_i$. The temporal encoder of our MAPLE projects the visible subset of embedding feature $g_i$ into latent space $\textit{Z}$, and the temporal decoder is responsible for reconstructing from the latent representation $z_i$ and the learnable mask tokens $M$.
However, different from classical autoencoders, our MAPLE implicitly reconstructs the original signal through the pseudo-label generated from the classification head, rather than reconstructing the original signal itself. We introduce the training process in detail as follows:

\textbf{Masking} the short-term global feature $g_i$ that is extracted from the Spatial Extractor (SE) modules is the first step of our framework. Like what ImageMAE does in~\cite{MAE:journals/corr/abs-2111-06377}, we directly discard a subset (e.g., 50\%) of the original short-term global feature $g_i$ with random sampling. 
The motivation of masking is to help the model efficiently understand the order of actions via reconstructing the complete action sequence from the mutilated one.

\textbf{Temporal Encoder (TE)} is a lightweight transformer that only contains several self-attention blocks in the second step of our MAPLE. We directly feed the masked short-term global features $g_i$ into the TE module without adding positional embedding, since its temporal positional embedding is already added when fed into the SE module. 

\textbf{Temporal Decoder (TD)} is also a lightweight transformer that is used to reconstruct the removed embedding feature $g_i$ in the third step of our MAPLE. Before feeding the latent representation $z_i$ into the TD modules, we first insert the shared and learnable mask token $M$ at the position of the original abandoned features and then add the new temporal positional embedding to the full set of sequences. Note that the shared mask without new temporal positional embedding cannot reconstruct the action information at different times.

\textbf{Reconstruction Target.} As shown in the final step of our MAPLE in Fig.~\ref{fig:teaser3} (b), the target of reconstruction is calculated with pseudo-label $P_i$ instead of the original feature $g_i$. We feed both the original feature $g_i$ and the reconstructed feature $r_i$ into the classification head to obtain their corresponding pseudo-label $P_i$ and $\hat{P}_i$. Note that the original pseudo-label $P_i$ is generated without backprop for maintaining the stability of the training process. Following this target, our unsupervised loss can be defined with the Kullback-Leibler divergence:
\begin{equation}
     L_u = L_{maple} =  \frac{1}{|D_u|} \sum_{x_i \in D_u} KL( f (P_{i} | x_{i}) || f (\hat{P}_i | x_{i}) ) ,
\end{equation}
where $KL$ is the function of Kullback-Leibler divergence, $|D_u|$ is the size of the unlabeled dataset, $f(\cdot)$ is the model, $P_i$ is the pseudo-label generated from the original feature $g_i$, $\hat{P}_i$ is the reconstructed pseudo-label generated from the reconstructed feature $r_i$. $Algorithm$~\ref{alg:algorithm1} and $Algorithm$~\ref{alg:algorithm2} present the training and inference process of our MAPLE, respectively.

Compare to reconstructing the original feature, reconstructing the pseudo-label not only improves the performance of classification but also makes the training stage more stable. We compare these two strategies in detail in section~\ref{subsec:Ablation Study}.

\begin{algorithm}[t]
\caption{The training process of our MAPLE.} \label{alg:algorithm1}
\begin{algorithmic}
\STATE
\STATE \textbf{Stage 1}: Pre-training with labeled dataset $D_l$ (corresponding to the left part of Fig.~\ref{fig:teaser3}). 

\STATE \textbf{Initialization}: the network parameters of DestFormer $\theta$; basic learning rate $\eta$; the labeled batch size $b_l$; the supervised cross-entropy loss $L_l$.
\STATE \textbf{repeat}
\INDSTATE t = 1 ... max iteration num:

\INDSTATE\hspace{\algorithmicindent} fetch mini-batch $d_l$ from $D_l$;

\INDSTATE\hspace{\algorithmicindent} compute loss $L_l$ on $d_l$;

\INDSTATE\hspace{\algorithmicindent} update $\theta^t = \theta^{t-1} - \eta \bigtriangledown L_l$.

\STATE \textbf{until} stable accuracy and loss in the validation set.

\STATE

\STATE \textbf{Stage 2}: Training of MAPLE with unlabeled dataset $D_u$ (corresponding to the left part of Fig.~\ref{fig:teaser3}).

\STATE \textbf{Initialization}: positive scalar weight $\alpha$ for unsupervised loss $L_u$ ; unlabeled batch size $b_u$, where $b_u \geq b_l$. 

\STATE \textbf{repeat}
\INDSTATE t = 1 ... max iteration num:

\INDSTATE\hspace{\algorithmicindent} fetch mini-batch $d_l$ from $D_l$ and $d_u$ from $D_u$;

\INDSTATE\hspace{\algorithmicindent} compute loss $L = L_l + \alpha L_u$ on $d_l$ and $d_u$;

\INDSTATE\hspace{\algorithmicindent} update $\theta^t = \theta^{t-1} - \eta \bigtriangledown L$.

\STATE \textbf{until} stable accuracy and loss in the validation set.

\end{algorithmic}

\end{algorithm}

\begin{algorithm}[t]
\caption{The inference process of our MAPLE.} \label{alg:algorithm2}
\begin{algorithmic}

\STATE
\STATE \textbf{Initialization}: the DestFormer model $f(\cdot)$ without the temporal decoder; the best-trained network parameters ${\theta}$.
\STATE \textbf{repeat}
\INDSTATE t = 1 ... final test batch:

\INDSTATE\hspace{\algorithmicindent} fetch mini-batch $d_t$ from test dataset $D_t$;

\INDSTATE\hspace{\algorithmicindent} calculate the accuracy on $d_t$;

\STATE \textbf{finished.} 

\STATE Calculate the accuracy on the whole test dataset $D_t$.

\STATE

\end{algorithmic}

\end{algorithm}

\section{Experiments}
To show the effectiveness of our DestFormer and MAPLE, we first evaluate the supervised-only performance and computational efficiency of our DestFormer. Then we compare our MAPLE with the semi-supervised baseline algorithms and further combine these leading algorithms with our MAPLE to obtain superior classification performance. At last, we investigate the choices of masking rate, the depth of temporal decoder, and the irreplaceability of pseudo-label.

\subsection{Dataset}

Our experiments are performed on three main human action recognition datasets: MSR-Action3D~\cite{MSRAction:conf/cvpr/LiZL10}, NTU RGB+D 60~\cite{NTU60:conf/cvpr/ShahroudyLNW16}, and NTU RGB+D 120~\cite{NTU120:journals/pami/LiuSPWDK20}.

\textbf{MSR-Action3D}~\cite{MSRAction:conf/cvpr/LiZL10} dataset captured with Kinect v1 depth camera, which contains 567 videos and 23k frames in total (270 videos for training and 297 videos for testing). This dataset contains twenty actions: high arm wave, horizontal arm wave, and so on.
For our semi-supervised point cloud action recognition, 7.5\%, 15.0\%, 22.5\%, 30.0\%, and 37.5\% of training videos of each action are selected for the labeled dataset $D_l$ and the rest for the unlabeled dataset $D_u$. More detailed information is available in the supplementary material.

\textbf{NTU RGB+D 60}~\cite{NTU60:conf/cvpr/ShahroudyLNW16} is a large dataset that was captured with Kinect v2 depth camera. It consists of 56K videos and 4M frames captured from 80 views and with 40 performers. Sixty action categories and two types of evaluation (i.e. cross-subject and cross-view) are defined in this dataset. In this paper, we evaluate our model with a cross-subject setting. For our semi-supervised point cloud action recognition task, 5\%, 10\%, 20\%, 30\%, and 40\% of training videos of each action are selected for the labeled dataset $D_l$.
 
\textbf{NTU RGB+D 120}~\cite{NTU120:journals/pami/LiuSPWDK20} is an extension of NTU RGB+D 60 and the largest dataset for human action recognition. It consists of 114K videos and 8M frames captured from 155 views and with 106 performers. The dataset captured by Kinect v2 depth camera has the modalities of RGB, Depth, 3DJoints, and IR. One hundred and twenty action categories and two types of evaluation (i.e. cross-subject and cross-setup) are defined on this dataset. To harmonize with the above dataset, we still evaluate our model with a cross-subject setting and select the same percentage of the labeled dataset $D_l$ as NTU RGB+D 60.

\subsection{Implementation Details and Approaches}
This subsection presents the training hyperparameters and implementation details of our DestFormer and MAPLE.

\textbf{Network Structure.} The DestFormer $f(\cdot)$ are introduced in Section~\ref{subsec:P4ST-Trans}. By default, The spatial scaling rate $\kappa$ of P4Conv is set to 2 and the spatial scaling rate $\zeta$ is set to 32. The spatial transformer is designed with 4 self-attention blocks and the temporal encoder is designed with only 3 self-attention blocks. Each black spatial transformer and temporal encoder contains 8 heads. As the Temporal Decoder used in MAPLE, it consists of 8 self-attention blocks to strengthen its ability for reconstruction.

\textbf{MAPLE Training.} In the whole process of training, the basic learning rate $\eta$ is set to 0.01. The warm-up strategy is used for the first 10 epochs with the initial $\eta = 10^{-6}$ and the decreased learning rate $\eta$ of the final 5 epochs is set to 0.001.
The mini-batch of $D_l$ and $D_u$ is set to 14 for the MSR-Action3D dataset, and 32 for NTU RGB+D 60 and NTU RGB+D 120. The masking ratio is set to 75\% for all datasets.
In Step 1 (Pre-training) of our training, the DestFormer is trained on the labeled dataset of MSR-Action3D, NTU RGB+D 60 and NTU RGB+D 120 with the epoch of 40, 20, and 20, respectively.
In Step 2 (Training of MAPLE), we set the positive scalar weight $\alpha$ as 0.5 for MSR-Action3D, and 0.2 for NTU RGB+D 60 and NTU RGB+D 120.

\begin{table}[t]
\centering
\setlength{\tabcolsep}{1mm}{
\begin{tabular}{l|l|ccccc} 
\hline
\multirow{2}{*}{Dataset}     & \multirow{2}{*}{Backbone} & \multicolumn{5}{c}{ Ratio of Labeled Data}       \\ 
\cline{3-7}
                             &                           & 7.5\%   & 15.0\%  & 22.5\%  & 30.0\%  & 37.5\%   \\ 
\hline 
\multirow{2}{*}{MSR3D~\cite{MSRAction:conf/cvpr/LiZL10}} & P4Transformer             & 61.95 & 77.10  & 80.47 & 83.16 & 85.85  \\
                             & DestFormer    & 62.96 & 77.44 & 81.14 & 83.84 & 86.53  \\ 
\hline \hline 
\multirow{2}{*}{Dataset}     & \multirow{2}{*}{Backbone} & \multicolumn{5}{c}{ Ratio of Labeled Data}       \\ 
\cline{3-7}
                             &                           & 5\%   & 10\%  & 20\%  & 30\%  & 40\%   \\ 
\hline
\multirow{2}{*}{NTU60~\cite{NTU60:conf/cvpr/ShahroudyLNW16}}      & P4Transformer             & 45.21 & 57.20  & 68.41 & 73.98 & 77.26  \\
                             & DestFormer    & 46.80  & 59.63 & 70.03 & 74.98 & 78.16  \\ 
\hline
\multirow{2}{*}{NTU120~\cite{NTU120:journals/pami/LiuSPWDK20}}     & P4Transformer            & 30.38 & 40.34 & 48.66 & 53.28 & 56.94  \\
                             & DestFormer    & 36.09 & 47.75 & 58.05 & 62.56 & 65.31  \\
\hline
\end{tabular}}
 \caption{Comparison of the supervised-only action recognition accuracy (\%) between P4Transformer~\cite{P4D:conf/cvpr/Fan0K21} and our DestFormer on three benchmark dataset.}
\label{tab:table1}
\end{table}

\begin{table}[t]
\centering
\setlength{\tabcolsep}{2mm}{
\begin{tabular}{l|c|c|c} 
\hline
Backbone & Depth & GFLOPs & Inference Time (ms) \\
\hline
P4Transformer~\cite{P4D:conf/cvpr/Fan0K21} & 5 & 85.6 & 865  \\
DestFormer  & 4+3  & 60.5 & 665   \\

\hline
\end{tabular}}
 \caption{Comparison of the time complexity (GFLOPs) and  average inference time (ms) between P4Transformer~\cite{P4D:conf/cvpr/Fan0K21} and Our DestFormer.}
\label{tab:table2}
\vspace{-5mm}
\end{table}

\textbf{Compared Approaches.} In the following section~\ref{Semi-Supervised Results}, we use leading semi-supervised learning algorithms that have proven to be generally effective as our compared approaches:

1) \textit{Supervised-only.} We train the model only with labeled dataset $D_l$. The performance of the best-trained model is used as the lower bounds of semi-supervised learning. 

2) \textit{Pseudo Labels~\cite{Pseudo:conf/icml/lee2013pseudo}.} 
The main idea is to further train the model with the pseudo hard labels of unlabeled data. This algorithm can be summarized in two steps as follows. First, we get the pre-training model with the supervised-only method, then we predict the pseudo hard labels of unlabeled data. Finally, the model can be retrained with these hard labels.

3) \textit{Virtual Adversarial Training (VAT)~\cite{VAT:journals/pami/MiyatoMKI19}.} It is inspired by adversarial learning and its regularization only needs unlabeled data. In the training process, it first adds small adversarial perturbation ${\epsilon}_{vat}$ to the unlabeled data for changing the final prediction, and then forces the model $f_{\theta}(\cdot)$ against this type of perturbation with the following consistency loss:

\begin{equation}
   L_{vat} =  \frac{1}{|D_u|} \sum_{x_i \in D_u} KL(f_{\theta}(x_i) ~ || ~ f_{\theta}(x_i+\triangle x_i)),
\end{equation}

\begin{equation}
    where~\triangle x_i = \mathop{\arg\max}\limits_{\delta~s.t.~ |\delta|_2 = {\epsilon}_{vat}}~KL(f_{\theta}(x_i) ~ || ~ f_{\theta}(x_i+\triangle x_i)).
\end{equation}

4) \textit{Conditional Entropy Minimization (EntMin)~\cite{EntMin:conf/nips/GrandvaletB04}.} This approach encourages the model to output the confident pseudo labels $y$ for unlabeled input data. In other words, the predictions $y$ closed to the one-hot vector are encouraged. The conditional entropy minimization loss can be defined as:

\begin{equation}
   L_{entmin} = \frac{1}{|D_u|} \sum_{x_i \in D_u} \sum_{y \in Y} - f_{\theta}(y|x_i) \log_{}{f_{\theta}(y|x_i)}.
\end{equation}

Note that the EntMin is almost not used alone for semi-supervised learning because the model can easily increase the weights of the classification head to generate a confident prediction. It always adopt with the VAT loss, i.e. $L_u = {\alpha}_{vat}L_{vat} + {\alpha}_{entmin}L_{entmin}$, where ${\alpha}_{vat}$ and ${\alpha}_{entmin}$ are the positive scalar weight for loss of VAT and EntMin, respectively.

\subsection{Supervised-only Performance}

To demonstrate the validity of our spatial-temporal backbone, this subsection compares the action recognition performance and computational efficiency of our DestFormer and the P4transformer model~\cite{P4D:conf/cvpr/Fan0K21} with the supervised-only setting on MSR-Action3D, NTU RGB+D 60 and NTU RGB+D 120 datasets. The action recognition accuracy (\%) of each backbone on three benchmark datasets is listed in Table~\ref{tab:table1}. The time complexity (GFLOPs) and  average inference time (ms) of each point cloud video are listed in Table~\ref{tab:table2}.

In Table~\ref{tab:table1}, we observe that our DestFormer has less annotation dependence and better classification performance in the supervised-only setting. Especially on the NTU RGB+D 120 dataset, our DestFormer model generally obtains a greater than 5.7\% increase in action recognition accuracy. 

In Table~\ref{tab:table2}, we notice that our DestFormer is more efficient in computational complexity, which obtain about 30\% and 23\% decrease for time complexity (GLOPs) and inference time (ms), respectively.

\subsection{Evaluation of Semi-supervised Methods}
\label{Semi-Supervised Results}

In this section, we first evaluate our MAPLE method by comparing it with leading semi-supervised methods (e.g. Pseudo Label, VAT, and EntMin) for semi-supervised point cloud action recognition on three mainstream datasets. Then we further combined our MAPLE method with those methods (VAT+EntMin+MAPLE) and obtain better performance for semi-supervised point cloud action recognition. Specifically, we use $L_{vat}$ and $L_{entmin}$ as unsupervised loss functions $L_u$ in the early training stage until the model has almost stabilized and then adopt the $L_{maple}$ as the unsupervised loss function. Please refer to the supplementary materials for the detailed training process of ``VAT+EntMin+MAPLE''.

\begin{table}[t]
\centering
\setlength{\tabcolsep}{1mm}{
\begin{tabular}{l|ccccc}
\hline
\multirow{2}{*}{Method} & \multicolumn{5}{c}{ Ratio of Labeled Data}       \\

\cline{2-6}
                        & 7.5\%  & 15.0\% & 22.5\% & 30.0\% & 37.5\%  \\
\hline
supervised-only         & 62.96 & 77.44 & 81.14 & 83.84 & 86.53  \\
\hline
Pseudo Label~\cite{Pseudo:conf/icml/lee2013pseudo}   & 68.01 & 80.64 & 81.65 & 85.19 & 88.05  \\
VAT~\cite{VAT:journals/pami/MiyatoMKI19}             & 66.92 & 80.47 & 81.14 & 85.19 & 86.53  \\
VAT + EntMin~\cite{EntMin:conf/nips/GrandvaletB04}   & 67.24 & 81.48 & 83.84 & 85.94 & 87.29  \\
\hline
MAPLE (Ours)                  & 72.04 & 82.15 & 84.85 & 87.04 & 89.40  \\
VAT+EntMin+MAPLE (Ours)        & \textbf{76.09} & \textbf{84.85} & \textbf{86.20} & \textbf{87.21} & \textbf{89.56} \\
\hline
\end{tabular}}
 \caption{Comparison of the results on MSR-Action3D.}
\label{tab:table3}
\vspace{-2mm}
\end{table}

\begin{table}[t]
\centering
\setlength{\tabcolsep}{1mm}{
\begin{tabular}{l|ccccc} 
\hline
\multirow{2}{*}{Method} & \multicolumn{5}{c}{ Ratio of Labeled Data}                                                    \\ 
\cline{2-6}
                        & 5\%          & 10\%           & 20\%           & 30\%   & 40\%            \\                         

\hline
supervised-only         & 46.80          & 59.63          & 70.03          & 74.98          & 78.16           \\ 
\hline
Pseudo Label~\cite{Pseudo:conf/icml/lee2013pseudo}            & 47.24          & 61.96          & 72.14          & 76.74          & 79.15           \\
VAT~\cite{VAT:journals/pami/MiyatoMKI19}                     & 46.80          & 59.95          & 70.92          & 75.77          & 78.47           \\
VAT + EntMin~\cite{EntMin:conf/nips/GrandvaletB04}            & 47.07          & 62.20          & 72.59          & 77.25          & 79.33           \\ 
\hline
MAPLE (Ours)                   & 48.78          & 60.61          & 71.05          & 75.72          & 78.61           \\
VAT+EntMin+MAPLE (Ours)        & \textbf{50.63} & \textbf{62.98} & \textbf{73.01} & \textbf{77.57} & \textbf{79.96}  \\
\hline
\end{tabular}}
\caption{Comparison of the results on NTU RGB+D 60.}
\label{tab:table4}
\vspace{-2mm}
\end{table}

\begin{figure}[t]
  \centering
  \vspace{0mm}
  \includegraphics[width=0.99\linewidth]{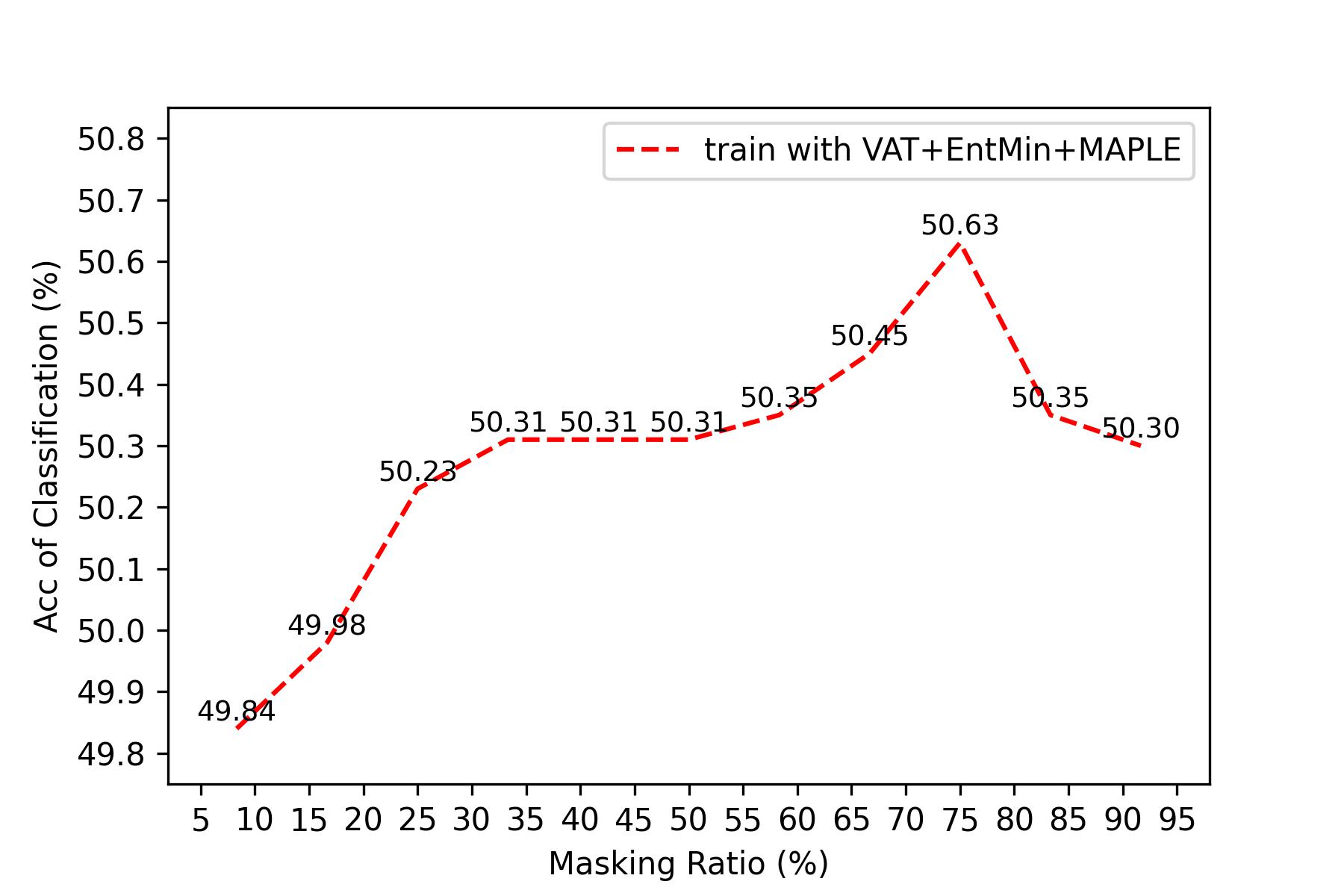}
  \caption{The accuracy of classification on NTU RGB+D 60 5\% labeled dataset with different masking ratios. The 75\% masking ratio of reconstruction achieves peak accuracy.}
  \label{fig:teaser4}
\end{figure}

\begin{figure*}[t]
  \centering
    \includegraphics[width=0.8\linewidth]{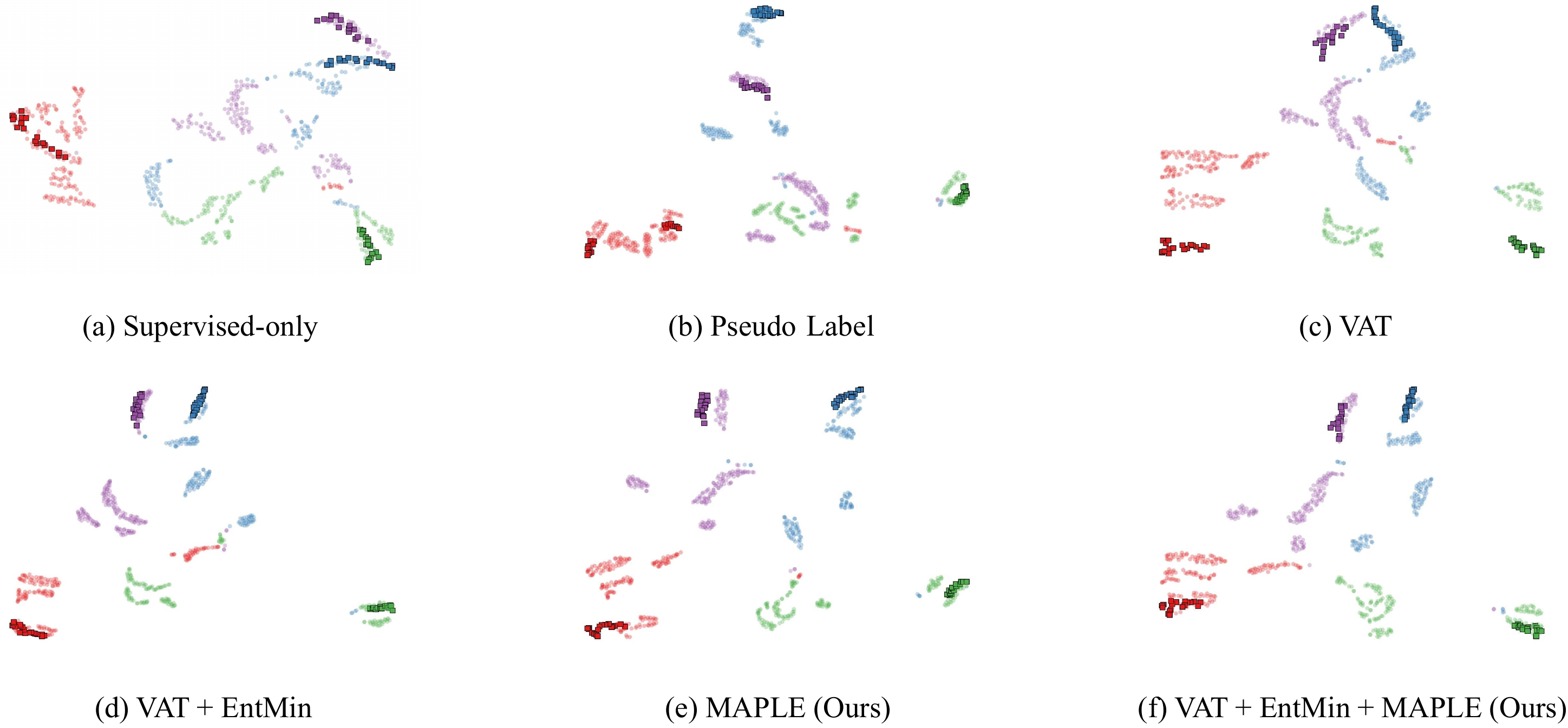}
  \caption{The t-SNE visualization of different approaches on the MSR-Action3D dataset. The squares with the black border indicate the labeled data, and other dots indicate the unlabeled ones. Note that different colors denote different classes.}
  \label{fig:teaser5}
\vspace{-3mm}
\end{figure*}

The results on MSR-Action3D and NTU RGB+D 60 datasets are shown in Tables~\ref{tab:table3} and ~\ref{tab:table4} respectively. The results on NTU RGB+D 120 are shown in the supplementary material.
From the semi-supervised results of each semi-supervised method, we can find that our MAPLE method is effective for semi-supervised learning and slightly outperforms the previous methods' performance under most settings. We also observe that our MAPLE method can be combined with other leading semi-supervised methods, and obtain significant performance increases under each setting. Especially on the 7.5\% labeled MSR-Action3D setting, it brings significant improvement
in action recognition performance (+8.08\% Acc).

\subsection{Ablation Study and Visualization}
\label{subsec:Ablation Study}
We investigate the effectiveness of our proposed MAPLE method on benchmark datasets in this section. We first analyze the influence of the masking ratio and the depth of the temporal decoder, then illustrate the importance of reconstruction with pseudo-label during the training process. At last, we show the feature distributions of each method to prove the effectiveness of our MAPLE method.

\textbf{Masking ratio.} Fig.~\ref{fig:teaser4} shows the accuracy of classification on NTU RGB+D 60 5\% labeled dataset with different masking ratios. The high masking ratio (75\%) of reconstruction achieves the peak of classification performance, which is as high as the masking ratio of Image MAE~\cite{MAE:journals/corr/abs-2111-06377}. This phenomenon is the exact opposite of the natural language processing field, whose best masking ratio is 15\% typically. However, this behavior verifies our hypothesis that most frames of the action sequence are redundant and we can reconstruct the complete action sequence from the small residual part of the sequence.

\textbf{Depth of Temporal Decoder.} We investigate the effect of decoder depth on the final classification performance and the result on NTU RGB+D 60 5\% labeled dataset with different depth of temporal decoder. For more detail, please refer to the supplementary materials.

\textbf{Reconstruction with Pseudo-Label.} To demonstrate the importance of implicit reconstruction through pseudo-label, we show the results with and without pseudo-label on the MSR-Action3D dataset in the supplementary materials.

\textbf{The t-SNE visualization.} 
To further explore the mechanism of MAPLE, we visualize the feature distributions of the labeled and unlabeled video sequences of the MSR-Action3D training dataset with t-SNE in Fig.~\ref{fig:teaser5}. 
From the t-SNE visualization, we can find that the model trained with only supervised action sequences is often hard to distinguish the decision boundaries of the unlabeled action sequence.
Although the benchmark methods (Pseudo Label, VAT, and EntMin) have better distributions, there are still some outliers that make the decision boundaries ambiguous.
Compared to previous approaches, our MAPLE and ``VAT+EntMin+MAPLE'' form tighter data clusters and clear decision boundaries which benefit from the semi-supervision learning with our masked autoencoder. 
To summarize, the visualization shows that combining semi-supervised learning with masked pseudo-labeling autoencoder is possible to learn numerous action concepts from unlabeled point cloud videos and improves the performance of action recognition.

\section{Conclusion}
In this paper, we present a Masked Pseudo-Labeling autoEncoder (\textbf{MAPLE}) framework with an effective transformer-based \textbf{De}coupled \textbf{s}patial-\textbf{t}emporal Trans\textbf{Former} (\textbf{DestFormer}) backbone to learn discriminative representations with much fewer annotations for the semi-supervised point cloud action recognition task. The MAPLE framework exploits the reconstruction of the masked features from the available frames to learn the numerous action concepts from unlabeled action sequences. Moreover, we combine our MAPLE with the classical semi-supervised methods to learn more generalizable features and establish the state-of-the-art performances of the semi-supervised point cloud action recognition task. We hope that our MAPLE framework can inspire the research of autoencoder on point cloud sequence in the future.

\begin{acks}
This research was supported by the National Key R\&D Program of China under Grant No. 2020AAA0103800.
\end{acks}

\newpage
\balance
\bibliographystyle{ACM-Reference-Format}
\bibliography{sample-base}


\begin{thebibliography}{40}


\ifx \showCODEN    \undefined \def \showCODEN     #1{\unskip}     \fi
\ifx \showDOI      \undefined \def \showDOI       #1{#1}\fi
\ifx \showISBNx    \undefined \def \showISBNx     #1{\unskip}     \fi
\ifx \showISBNxiii \undefined \def \showISBNxiii  #1{\unskip}     \fi
\ifx \showISSN     \undefined \def \showISSN      #1{\unskip}     \fi
\ifx \showLCCN     \undefined \def \showLCCN      #1{\unskip}     \fi
\ifx \shownote     \undefined \def \shownote      #1{#1}          \fi
\ifx \showarticletitle \undefined \def \showarticletitle #1{#1}   \fi
\ifx \showURL      \undefined \def \showURL       {\relax}        \fi
\providecommand\bibfield[2]{#2}
\providecommand\bibinfo[2]{#2}
\providecommand\natexlab[1]{#1}
\providecommand\showeprint[2][]{arXiv:#2}

\bibitem[Beyer et~al\mbox{.}(2019)]%
        {S4L:conf/iccv/BeyerZOK19}
\bibfield{author}{\bibinfo{person}{Lucas Beyer}, \bibinfo{person}{Xiaohua
  Zhai}, \bibinfo{person}{Avital Oliver}, {and} \bibinfo{person}{Alexander
  Kolesnikov}.} \bibinfo{year}{2019}\natexlab{}.
\newblock \showarticletitle{{S4L:} Self-Supervised Semi-Supervised Learning}.
  In \bibinfo{booktitle}{\emph{{ICCV}}}. \bibinfo{pages}{1476--1485}.
\newblock


\bibitem[Chapelle et~al\mbox{.}(2006)]%
        {SSL:books/mit/06/CSZ2006}
\bibfield{editor}{\bibinfo{person}{Olivier Chapelle}, \bibinfo{person}{Bernhard
  Sch{\"{o}}lkopf}, {and} \bibinfo{person}{Alexander Zien}} (Eds.).
  \bibinfo{year}{2006}\natexlab{}.
\newblock \bibinfo{booktitle}{\emph{Semi-Supervised Learning}}.
\newblock


\bibitem[Chen et~al\mbox{.}(2020b)]%
        {SimCLR:conf/icml/ChenK0H20}
\bibfield{author}{\bibinfo{person}{Ting Chen}, \bibinfo{person}{Simon
  Kornblith}, \bibinfo{person}{Mohammad Norouzi}, {and}
  \bibinfo{person}{Geoffrey~E. Hinton}.} \bibinfo{year}{2020}\natexlab{b}.
\newblock \showarticletitle{A Simple Framework for Contrastive Learning of
  Visual Representations}. In \bibinfo{booktitle}{\emph{{ICML}}}.
  \bibinfo{pages}{1597--1607}.
\newblock


\bibitem[Chen et~al\mbox{.}(2020a)]%
        {MoCo2:journals/corr/abs-2003-04297}
\bibfield{author}{\bibinfo{person}{Xinlei Chen}, \bibinfo{person}{Haoqi Fan},
  \bibinfo{person}{Ross~B. Girshick}, {and} \bibinfo{person}{Kaiming He}.}
  \bibinfo{year}{2020}\natexlab{a}.
\newblock \showarticletitle{Improved Baselines with Momentum Contrastive
  Learning}.
\newblock \bibinfo{journal}{\emph{{CoRR}}} (\bibinfo{year}{2020}).
\newblock


\bibitem[Devlin et~al\mbox{.}(2019)]%
        {BERT:conf/naacl/DevlinCLT19}
\bibfield{author}{\bibinfo{person}{Jacob Devlin}, \bibinfo{person}{Ming{-}Wei
  Chang}, \bibinfo{person}{Kenton Lee}, {and} \bibinfo{person}{Kristina
  Toutanova}.} \bibinfo{year}{2019}\natexlab{}.
\newblock \showarticletitle{{BERT:} Pre-training of Deep Bidirectional
  Transformers for Language Understanding}. In
  \bibinfo{booktitle}{\emph{{NAACL-HLT}}}. \bibinfo{pages}{4171--4186}.
\newblock


\bibitem[Fan et~al\mbox{.}(2021)]%
        {P4D:conf/cvpr/Fan0K21}
\bibfield{author}{\bibinfo{person}{Hehe Fan}, \bibinfo{person}{Yi Yang}, {and}
  \bibinfo{person}{Mohan~S. Kankanhalli}.} \bibinfo{year}{2021}\natexlab{}.
\newblock \showarticletitle{Point 4D Transformer Networks for Spatio-Temporal
  Modeling in Point Cloud Videos}. In \bibinfo{booktitle}{\emph{{CVPR}}}.
  \bibinfo{pages}{14204--14213}.
\newblock


\bibitem[Grandvalet and Bengio(2004)]%
        {EntMin:conf/nips/GrandvaletB04}
\bibfield{author}{\bibinfo{person}{Yves Grandvalet} {and}
  \bibinfo{person}{Yoshua Bengio}.} \bibinfo{year}{2004}\natexlab{}.
\newblock \showarticletitle{Semi-supervised Learning by Entropy Minimization}.
  In \bibinfo{booktitle}{\emph{{NeurIPS}}}. \bibinfo{pages}{529--536}.
\newblock


\bibitem[Grill et~al\mbox{.}(2020)]%
        {BYOL:conf/nips/GrillSATRBDPGAP20}
\bibfield{author}{\bibinfo{person}{Jean{-}Bastien Grill},
  \bibinfo{person}{Florian Strub}, \bibinfo{person}{Florent Altch{\'{e}}},
  \bibinfo{person}{Corentin Tallec}, \bibinfo{person}{Pierre~H. Richemond},
  \bibinfo{person}{Elena Buchatskaya}, \bibinfo{person}{Carl Doersch},
  \bibinfo{person}{Bernardo~{\'{A}}vila Pires}, \bibinfo{person}{Zhaohan Guo},
  \bibinfo{person}{Mohammad~Gheshlaghi Azar}, \bibinfo{person}{Bilal Piot},
  \bibinfo{person}{Koray Kavukcuoglu}, \bibinfo{person}{R{\'{e}}mi Munos},
  {and} \bibinfo{person}{Michal Valko}.} \bibinfo{year}{2020}\natexlab{}.
\newblock \showarticletitle{Bootstrap Your Own Latent - {A} New Approach to
  Self-Supervised Learning}. In \bibinfo{booktitle}{\emph{NeurIPS}}.
  \bibinfo{pages}{21271--21284}.
\newblock


\bibitem[He et~al\mbox{.}(2021)]%
        {MAE:journals/corr/abs-2111-06377}
\bibfield{author}{\bibinfo{person}{Kaiming He}, \bibinfo{person}{Xinlei Chen},
  \bibinfo{person}{Saining Xie}, \bibinfo{person}{Yanghao Li},
  \bibinfo{person}{Piotr Doll{\'{a}}r}, {and} \bibinfo{person}{Ross~B.
  Girshick}.} \bibinfo{year}{2021}\natexlab{}.
\newblock \showarticletitle{Masked Autoencoders Are Scalable Vision Learners}.
\newblock \bibinfo{journal}{\emph{{CoRR}}} (\bibinfo{year}{2021}).
\newblock


\bibitem[He et~al\mbox{.}(2020)]%
        {MoCo:conf/cvpr/He0WXG20}
\bibfield{author}{\bibinfo{person}{Kaiming He}, \bibinfo{person}{Haoqi Fan},
  \bibinfo{person}{Yuxin Wu}, \bibinfo{person}{Saining Xie}, {and}
  \bibinfo{person}{Ross~B. Girshick}.} \bibinfo{year}{2020}\natexlab{}.
\newblock \showarticletitle{Momentum Contrast for Unsupervised Visual
  Representation Learning}. In \bibinfo{booktitle}{\emph{{CVPR}}}.
  \bibinfo{pages}{9726--9735}.
\newblock


\bibitem[Hinton et~al\mbox{.}(2015)]%
        {Distillation:journals/corr/HintonVD15}
\bibfield{author}{\bibinfo{person}{Geoffrey~E. Hinton}, \bibinfo{person}{Oriol
  Vinyals}, {and} \bibinfo{person}{Jeffrey Dean}.}
  \bibinfo{year}{2015}\natexlab{}.
\newblock \showarticletitle{Distilling the Knowledge in a Neural Network}.
\newblock \bibinfo{journal}{\emph{{CoRR}}} (\bibinfo{year}{2015}).
\newblock


\bibitem[Jebara et~al\mbox{.}(2009)]%
        {SSL2009:conf/icml/JebaraWC09}
\bibfield{author}{\bibinfo{person}{Tony Jebara}, \bibinfo{person}{Jun Wang},
  {and} \bibinfo{person}{Shih{-}Fu Chang}.} \bibinfo{year}{2009}\natexlab{}.
\newblock \showarticletitle{Graph construction and \emph{b}-matching for
  semi-supervised learning}. In \bibinfo{booktitle}{\emph{{ICML}}}.
  \bibinfo{pages}{441--448}.
\newblock


\bibitem[Lee et~al\mbox{.}(2013)]%
        {Pseudo:conf/icml/lee2013pseudo}
\bibfield{author}{\bibinfo{person}{Dong-Hyun Lee} {et~al\mbox{.}}}
  \bibinfo{year}{2013}\natexlab{}.
\newblock \showarticletitle{Pseudo-label: The simple and efficient
  semi-supervised learning method for deep neural networks}. In
  \bibinfo{booktitle}{\emph{Workshop on challenges in representation learning,
  ICML}}. \bibinfo{pages}{896}.
\newblock


\bibitem[Li et~al\mbox{.}(2010)]%
        {MSRAction:conf/cvpr/LiZL10}
\bibfield{author}{\bibinfo{person}{Wanqing Li}, \bibinfo{person}{Zhengyou
  Zhang}, {and} \bibinfo{person}{Zicheng Liu}.}
  \bibinfo{year}{2010}\natexlab{}.
\newblock \showarticletitle{Action recognition based on a bag of 3D points}. In
  \bibinfo{booktitle}{\emph{{CVPR} Workshops}}. \bibinfo{pages}{9--14}.
\newblock


\bibitem[Liu et~al\mbox{.}(2020a)]%
        {NTU120:journals/pami/LiuSPWDK20}
\bibfield{author}{\bibinfo{person}{Jun Liu}, \bibinfo{person}{Amir Shahroudy},
  \bibinfo{person}{Mauricio Perez}, \bibinfo{person}{Gang Wang},
  \bibinfo{person}{Ling{-}Yu Duan}, {and} \bibinfo{person}{Alex~C. Kot}.}
  \bibinfo{year}{2020}\natexlab{a}.
\newblock \showarticletitle{{NTU} {RGB+D} 120: {A} Large-Scale Benchmark for 3D
  Human Activity Understanding}.
\newblock \bibinfo{journal}{\emph{{IEEE} Trans. Pattern Anal. Mach. Intell.}}
  (\bibinfo{year}{2020}), \bibinfo{pages}{2684--2701}.
\newblock


\bibitem[Liu et~al\mbox{.}(2018)]%
        {l:conf/aaai/LiuLGTM18}
\bibfield{author}{\bibinfo{person}{Kun Liu}, \bibinfo{person}{Wu Liu},
  \bibinfo{person}{Chuang Gan}, \bibinfo{person}{Mingkui Tan}, {and}
  \bibinfo{person}{Huadong Ma}.} \bibinfo{year}{2018}\natexlab{}.
\newblock \showarticletitle{{T-C3D:} Temporal Convolutional 3D Network for
  Real-Time Action Recognition}. In \bibinfo{booktitle}{\emph{{AAAI}}}.
  \bibinfo{pages}{7138--7145}.
\newblock


\bibitem[Liu et~al\mbox{.}(2022)]%
        {l:liu2022recent}
\bibfield{author}{\bibinfo{person}{Wu Liu}, \bibinfo{person}{Qian Bao},
  \bibinfo{person}{Yu Sun}, {and} \bibinfo{person}{Mei Tao}.}
  \bibinfo{year}{2022}\natexlab{}.
\newblock \showarticletitle{Recent Advances of Monocular 2D and 3D Human Pose
  Estimation: A Deep Learning Perspective}.
\newblock \bibinfo{journal}{\emph{ACM Computing Surveys (CSUR)}}
  (\bibinfo{year}{2022}).
\newblock


\bibitem[Liu et~al\mbox{.}(2010)]%
        {SSL2010:conf/icml/LiuHC10}
\bibfield{author}{\bibinfo{person}{Wei Liu}, \bibinfo{person}{Junfeng He},
  {and} \bibinfo{person}{Shih{-}Fu Chang}.} \bibinfo{year}{2010}\natexlab{}.
\newblock \showarticletitle{Large Graph Construction for Scalable
  Semi-Supervised Learning}. In \bibinfo{booktitle}{\emph{{ICML}}}.
  \bibinfo{pages}{679--686}.
\newblock


\bibitem[Liu et~al\mbox{.}(2012)]%
        {SSL2012:journals/pieee/LiuWC12}
\bibfield{author}{\bibinfo{person}{Wei Liu}, \bibinfo{person}{Jun Wang}, {and}
  \bibinfo{person}{Shih{-}Fu Chang}.} \bibinfo{year}{2012}\natexlab{}.
\newblock \showarticletitle{Robust and Scalable Graph-Based Semisupervised
  Learning}.
\newblock \bibinfo{journal}{\emph{Proc. {IEEE}}} (\bibinfo{year}{2012}),
  \bibinfo{pages}{2624--2638}.
\newblock


\bibitem[Liu et~al\mbox{.}(2019)]%
        {MeteorNet:conf/iccv/LiuYB19}
\bibfield{author}{\bibinfo{person}{Xingyu Liu}, \bibinfo{person}{Mengyuan Yan},
  {and} \bibinfo{person}{Jeannette Bohg}.} \bibinfo{year}{2019}\natexlab{}.
\newblock \showarticletitle{MeteorNet: Deep Learning on Dynamic 3D Point Cloud
  Sequences}. In \bibinfo{booktitle}{\emph{{ICCV}}}.
  \bibinfo{pages}{9245--9254}.
\newblock


\bibitem[Liu et~al\mbox{.}(2020b)]%
        {ssl2020:journals/corr/abs-2006-08218}
\bibfield{author}{\bibinfo{person}{Xiao Liu}, \bibinfo{person}{Fanjin Zhang},
  \bibinfo{person}{Zhenyu Hou}, \bibinfo{person}{Zhaoyu Wang},
  \bibinfo{person}{Li Mian}, \bibinfo{person}{Jing Zhang}, {and}
  \bibinfo{person}{Jie Tang}.} \bibinfo{year}{2020}\natexlab{b}.
\newblock \showarticletitle{Self-supervised Learning: Generative or
  Contrastive}.
\newblock \bibinfo{journal}{\emph{{CoRR}}} (\bibinfo{year}{2020}).
\newblock


\bibitem[Liu et~al\mbox{.}(2021)]%
        {Swin:conf/iccv/LiuL00W0LG21}
\bibfield{author}{\bibinfo{person}{Ze Liu}, \bibinfo{person}{Yutong Lin},
  \bibinfo{person}{Yue Cao}, \bibinfo{person}{Han Hu}, \bibinfo{person}{Yixuan
  Wei}, \bibinfo{person}{Zheng Zhang}, \bibinfo{person}{Stephen Lin}, {and}
  \bibinfo{person}{Baining Guo}.} \bibinfo{year}{2021}\natexlab{}.
\newblock \showarticletitle{Swin Transformer: Hierarchical Vision Transformer
  using Shifted Windows}. In \bibinfo{booktitle}{\emph{{ICCV}}}.
  \bibinfo{pages}{9992--10002}.
\newblock


\bibitem[Martin-Martin et~al\mbox{.}(2021)]%
        {JRDB:journals/tpami/martin2021jrdb}
\bibfield{author}{\bibinfo{person}{Roberto Martin-Martin},
  \bibinfo{person}{Mihir Patel}, \bibinfo{person}{Hamid Rezatofighi},
  \bibinfo{person}{Abhijeet Shenoi}, \bibinfo{person}{JunYoung Gwak},
  \bibinfo{person}{Eric Frankel}, \bibinfo{person}{Amir Sadeghian}, {and}
  \bibinfo{person}{Silvio Savarese}.} \bibinfo{year}{2021}\natexlab{}.
\newblock \showarticletitle{JRDB: A dataset and benchmark of egocentric robot
  visual perception of humans in built environments}.
\newblock \bibinfo{journal}{\emph{{TPAMI}}} (\bibinfo{year}{2021}).
\newblock


\bibitem[Miyato et~al\mbox{.}(2019)]%
        {VAT:journals/pami/MiyatoMKI19}
\bibfield{author}{\bibinfo{person}{Takeru Miyato}, \bibinfo{person}{Shin{-}ichi
  Maeda}, \bibinfo{person}{Masanori Koyama}, {and} \bibinfo{person}{Shin
  Ishii}.} \bibinfo{year}{2019}\natexlab{}.
\newblock \showarticletitle{Virtual Adversarial Training: {A} Regularization
  Method for Supervised and Semi-Supervised Learning}.
\newblock \bibinfo{journal}{\emph{{IEEE} Trans. Pattern Anal. Mach. Intell.}}
  (\bibinfo{year}{2019}), \bibinfo{pages}{1979--1993}.
\newblock


\bibitem[Qi et~al\mbox{.}(2017a)]%
        {pointnet:conf/cvpr/QiSMG17}
\bibfield{author}{\bibinfo{person}{Charles~Ruizhongtai Qi},
  \bibinfo{person}{Hao Su}, \bibinfo{person}{Kaichun Mo}, {and}
  \bibinfo{person}{Leonidas~J. Guibas}.} \bibinfo{year}{2017}\natexlab{a}.
\newblock \showarticletitle{PointNet: Deep Learning on Point Sets for 3D
  Classification and Segmentation}. In \bibinfo{booktitle}{\emph{{CVPR}}}.
  \bibinfo{pages}{77--85}.
\newblock


\bibitem[Qi et~al\mbox{.}(2017b)]%
        {pointnet2:conf/nips/QiYSG17}
\bibfield{author}{\bibinfo{person}{Charles~Ruizhongtai Qi}, \bibinfo{person}{Li
  Yi}, \bibinfo{person}{Hao Su}, {and} \bibinfo{person}{Leonidas~J. Guibas}.}
  \bibinfo{year}{2017}\natexlab{b}.
\newblock \showarticletitle{PointNet++: Deep Hierarchical Feature Learning on
  Point Sets in a Metric Space}. In \bibinfo{booktitle}{\emph{{NeurIPS}}}.
  \bibinfo{pages}{5099--5108}.
\newblock


\bibitem[Radford et~al\mbox{.}(2018)]%
        {GPT:journals/corr/radford2018improving}
\bibfield{author}{\bibinfo{person}{Alec Radford}, \bibinfo{person}{Karthik
  Narasimhan}, \bibinfo{person}{Tim Salimans}, {and} \bibinfo{person}{Ilya
  Sutskever}.} \bibinfo{year}{2018}\natexlab{}.
\newblock \showarticletitle{Improving language understanding by generative
  pre-training}.
\newblock \bibinfo{journal}{\emph{{CoRR}}} (\bibinfo{year}{2018}).
\newblock


\bibitem[Razavi et~al\mbox{.}(2019)]%
        {vqvae:conf/nips/RazaviOV19}
\bibfield{author}{\bibinfo{person}{Ali Razavi}, \bibinfo{person}{A{\"{a}}ron
  van~den Oord}, {and} \bibinfo{person}{Oriol Vinyals}.}
  \bibinfo{year}{2019}\natexlab{}.
\newblock \showarticletitle{Generating Diverse High-Fidelity Images with
  {VQ-VAE-2}}. In \bibinfo{booktitle}{\emph{NeurIPS}}.
  \bibinfo{pages}{14837--14847}.
\newblock


\bibitem[Shahroudy et~al\mbox{.}(2016)]%
        {NTU60:conf/cvpr/ShahroudyLNW16}
\bibfield{author}{\bibinfo{person}{Amir Shahroudy}, \bibinfo{person}{Jun Liu},
  \bibinfo{person}{Tian{-}Tsong Ng}, {and} \bibinfo{person}{Gang Wang}.}
  \bibinfo{year}{2016}\natexlab{}.
\newblock \showarticletitle{{NTU} {RGB+D:} {A} Large Scale Dataset for 3D Human
  Activity Analysis}. In \bibinfo{booktitle}{\emph{{CVPR}}}.
  \bibinfo{pages}{1010--1019}.
\newblock


\bibitem[Sheikhpour et~al\mbox{.}(2017)]%
        {SSL2017:journals/pr/SheikhpourSGC17}
\bibfield{author}{\bibinfo{person}{Razieh Sheikhpour},
  \bibinfo{person}{Mehdi~Agha Sarram}, \bibinfo{person}{Sajjad Gharaghani},
  {and} \bibinfo{person}{Mohammad Ali~Zare Chahooki}.}
  \bibinfo{year}{2017}\natexlab{}.
\newblock \showarticletitle{A Survey on semi-supervised feature selection
  methods}.
\newblock \bibinfo{journal}{\emph{Pattern Recognit.}} (\bibinfo{year}{2017}),
  \bibinfo{pages}{141--158}.
\newblock


\bibitem[Subramanya and Talukdar(2014)]%
        {SSL2014:series/synthesis/2014Subramanya}
\bibfield{author}{\bibinfo{person}{Amarnag Subramanya} {and}
  \bibinfo{person}{Partha~Pratim Talukdar}.} \bibinfo{year}{2014}\natexlab{}.
\newblock \bibinfo{booktitle}{\emph{Graph-Based Semi-Supervised Learning}}.
\newblock \bibinfo{publisher}{Morgan {\&} Claypool Publishers}.
\newblock


\bibitem[Sun et~al\mbox{.}(2022)]%
        {l:conf/cvpr/SunLBFMB22}
\bibfield{author}{\bibinfo{person}{Yu Sun}, \bibinfo{person}{Wu Liu},
  \bibinfo{person}{Qian Bao}, \bibinfo{person}{Yili Fu}, \bibinfo{person}{Tao
  Mei}, {and} \bibinfo{person}{Michael~J. Black}.}
  \bibinfo{year}{2022}\natexlab{}.
\newblock \showarticletitle{Putting People in their Place: Monocular Regression
  of 3D People in Depth}.
\newblock  (\bibinfo{year}{2022}), \bibinfo{pages}{13243--13252}.
\newblock


\bibitem[Triguero et~al\mbox{.}(2015)]%
        {SSL2015:journals/kais/TrigueroGH15}
\bibfield{author}{\bibinfo{person}{Isaac Triguero}, \bibinfo{person}{Salvador
  Garc{\'{\i}}a}, {and} \bibinfo{person}{Francisco Herrera}.}
  \bibinfo{year}{2015}\natexlab{}.
\newblock \showarticletitle{Self-labeled techniques for semi-supervised
  learning: taxonomy, software and empirical study}.
\newblock \bibinfo{journal}{\emph{Knowl. Inf. Syst.}} (\bibinfo{year}{2015}),
  \bibinfo{pages}{245--284}.
\newblock


\bibitem[van~den Oord et~al\mbox{.}(2016)]%
        {PixelCNN:conf/icml/OordKK16}
\bibfield{author}{\bibinfo{person}{A{\"{a}}ron van~den Oord},
  \bibinfo{person}{Nal Kalchbrenner}, {and} \bibinfo{person}{Koray
  Kavukcuoglu}.} \bibinfo{year}{2016}\natexlab{}.
\newblock \showarticletitle{Pixel Recurrent Neural Networks}. In
  \bibinfo{booktitle}{\emph{{ICML}}}. \bibinfo{pages}{1747--1756}.
\newblock


\bibitem[Vaswani et~al\mbox{.}(2017)]%
        {Transformer:conf/nips/VaswaniSPUJGKP17}
\bibfield{author}{\bibinfo{person}{Ashish Vaswani}, \bibinfo{person}{Noam
  Shazeer}, \bibinfo{person}{Niki Parmar}, \bibinfo{person}{Jakob Uszkoreit},
  \bibinfo{person}{Llion Jones}, \bibinfo{person}{Aidan~N. Gomez},
  \bibinfo{person}{Lukasz Kaiser}, {and} \bibinfo{person}{Illia Polosukhin}.}
  \bibinfo{year}{2017}\natexlab{}.
\newblock \showarticletitle{Attention is All you Need}. In
  \bibinfo{booktitle}{\emph{{NeurIPS}}}. \bibinfo{pages}{5998--6008}.
\newblock


\bibitem[Wang et~al\mbox{.}(2020)]%
        {3DV:conf/cvpr/Wang0XJCZY20}
\bibfield{author}{\bibinfo{person}{Yancheng Wang}, \bibinfo{person}{Yang Xiao},
  \bibinfo{person}{Fu Xiong}, \bibinfo{person}{Wenxiang Jiang},
  \bibinfo{person}{Zhiguo Cao}, \bibinfo{person}{Joey~Tianyi Zhou}, {and}
  \bibinfo{person}{Junsong Yuan}.} \bibinfo{year}{2020}\natexlab{}.
\newblock \showarticletitle{3DV: 3D Dynamic Voxel for Action Recognition in
  Depth Video}. In \bibinfo{booktitle}{\emph{{CVPR}}}.
  \bibinfo{pages}{508--517}.
\newblock


\bibitem[Wei et~al\mbox{.}(2022)]%
        {PST2:conf/wacv/WeiLXKG22}
\bibfield{author}{\bibinfo{person}{Yimin Wei}, \bibinfo{person}{Hao Liu},
  \bibinfo{person}{Tingting Xie}, \bibinfo{person}{Qiuhong Ke}, {and}
  \bibinfo{person}{Yulan Guo}.} \bibinfo{year}{2022}\natexlab{}.
\newblock \showarticletitle{Spatial-Temporal Transformer for 3D Point Cloud
  Sequences}. In \bibinfo{booktitle}{\emph{{WACV}}}. \bibinfo{pages}{657--666}.
\newblock


\bibitem[X.(2008)]%
        {SSL:books/mit/08/Zhu2008}
\bibfield{editor}{\bibinfo{person}{Zhu X.}} (Ed.).
  \bibinfo{year}{2008}\natexlab{}.
\newblock \bibinfo{booktitle}{\emph{Semi-supervised learning literature survey:
  Department of Computer Sciences}}.
\newblock


\bibitem[Zhao et~al\mbox{.}(2021)]%
        {captity:journals/corr/abs-2108-13002}
\bibfield{author}{\bibinfo{person}{Yucheng Zhao}, \bibinfo{person}{Guangting
  Wang}, \bibinfo{person}{Chuanxin Tang}, \bibinfo{person}{Chong Luo},
  \bibinfo{person}{Wenjun Zeng}, {and} \bibinfo{person}{Zheng{-}Jun Zha}.}
  \bibinfo{year}{2021}\natexlab{}.
\newblock \showarticletitle{A Battle of Network Structures: An Empirical Study
  of CNN, Transformer, and {MLP}}.
\newblock \bibinfo{journal}{\emph{{CoRR}}} (\bibinfo{year}{2021}).
\newblock


\bibitem[Zheng et~al\mbox{.}(2022)]%
        {zheng2022gait3d}
\bibfield{author}{\bibinfo{person}{Jinkai Zheng}, \bibinfo{person}{Xinchen
  Liu}, \bibinfo{person}{Wu Liu}, \bibinfo{person}{Lingxiao He},
  \bibinfo{person}{Chenggang Yan}, {and} \bibinfo{person}{Tao Mei}.}
  \bibinfo{year}{2022}\natexlab{}.
\newblock \showarticletitle{Gait Recognition in the Wild with Dense 3D
  Representations and A Benchmark}. In \bibinfo{booktitle}{\emph{CVPR}}.
  \bibinfo{pages}{20228--20237}.
\newblock


\end{thebibliography}

\newpage
\appendix

\section{Additional Experimental Results}
\label{sec:intro}
In this supplementary material, we show more details about the semi-supervised datasets and our MAPLE algorithm.


\begin{table}[b]
\centering
\setlength{\tabcolsep}{1mm}{
\begin{tabular}{l|l|ccccc} 
\hline
\multirow{2}{*}{Dataset} & \multirow{2}{*}{Division} & \multicolumn{5}{c}{Ratio of Labeled Data}  \\ 
\cline{3-7}
                         &                           & 7.5\% & 15.0\% & 22.5\% & 30.0\% & 37.5\%  \\ 
\hline
\multirow{2}{*}{MSR3D}     & Labeled          & 20    & 40     & 60     & 80     & 100     \\
                         & Unlabeled        & 250   & 230    & 210    & 190    & 170     \\
\hline \hline 
\multirow{2}{*}{Dataset} & \multirow{2}{*}{Division} & \multicolumn{5}{c}{Ratio of Labeled Data}  \\ 
\cline{3-7}
                         &                           & 5.0\% & 10.0\% & 20.0\% & 30.0\% & 40.0\%  \\ 
\hline
\multirow{2}{*}{NTU60}   & Labeled          & 1980  & 3960   & 7920   & 11880  & 15840   \\
                         & Unlabeled        & 38340 & 36360  & 32400  & 28440  & 24480   \\
\hline
\multirow{2}{*}{NTU120}  & Labeled          & 3120  & 6358   & 12416  & 17846  & 22810   \\
                         & Unlabeled        & 60240 & 57002  & 50944  & 45514  & 40550   \\
\hline
\end{tabular}}

\caption{The division of the semi-supervised datasets.}
\label{supp_tab:table1}
\end{table}

\begin{table}[b]
\centering
\setlength{\tabcolsep}{1mm}{
\begin{tabular}{l|ccccc} 
\hline
\multirow{2}{*}{Method} & \multicolumn{5}{c}{ Ratio of Labeled Data}                                                    \\ 
\cline{2-6}
                        & 5\%          & 10\%           & 20\%           & 30\%   & 40\%            \\                         

\hline
Supervised-only         & 36.09          & 47.75          & 58.05          & 62.56         & 65.31          \\ 
\hline
Pseudo Label            & 36.18          & 48.25          & 58.33          & 62.85          & 65.55           \\
VAT                    & 35.90         & 47.74          & 58.29          & 62.56          & 65.75           \\
VAT + EntMin            & 36.02         & 48.2          & 58.42          & 62.70          & 66.88           \\ 
\hline
MAPLE (Ours)                   & \textbf{37.15}         & 48.56          & 58.59          & 63.18          & 65.84           \\
VAT+EntMin+MAPLE (Ours)        & 36.91 & \textbf{48.80} & \textbf{59.25} & \textbf{64.02} & \textbf{67.08}  \\
\hline
\end{tabular}}
\caption{Comparison of the results on the NTU120 dataset.}
\label{supp_tab:table2}
\end{table}

\begin{table}[b]
\centering
\setlength{\tabcolsep}{1mm}{
\begin{tabular}{l|ccccc} 
\hline
\multirow{2}{*}{Method} & \multicolumn{5}{c}{ Ratio of Labeled Data}                                                    \\ 
\cline{2-6}
                        & 7.5\%          & 15.0\%           & 22.5\%           & 30.0\%   & 37.5\%            \\                         

\hline

MAPLE w/o pseudo-label                   & 69.19        & 81.32          & 84.51          & 86.87          & 86.53           \\
MAPLE with pseudo-label       & \textbf{72.04} & \textbf{82.15} & \textbf{84.85} & \textbf{87.04} & \textbf{89.40}  \\
\hline
\end{tabular}}
\caption{The results of MAPLE with and w/o pseudo-label on the MSR-Action3D dataset}
\label{supp_tab:table3}
\end{table}

\begin{algorithm}[b]
\caption{The training process of our VAT+EntMin+MAPLE.} \label{supp_alg:algorithm1}
\begin{algorithmic}
\STATE
\STATE \textbf{Stage 1}: Pre-training with labeled dataset $D_l$. 

\STATE \textbf{Initialization}: the network parameters of DestFormer $\theta$; basic learning rate $\eta$; the labeled batch size $b_l$; the supervised cross-entropy loss $L_l$.
\STATE \textbf{repeat}
\INDSTATE t = 1 ... max iteration num:

\INDSTATE\hspace{\algorithmicindent} fetch mini-batch $d_l$ from $D_l$;

\INDSTATE\hspace{\algorithmicindent} compute loss $L_l$ on $d_l$;

\INDSTATE\hspace{\algorithmicindent} update $\theta^t = \theta^{t-1} - \eta \bigtriangledown L_l$.

\STATE \textbf{until} stable accuracy and loss in the validation set.

\STATE

\STATE \textbf{Stage 2}: Training of VAT+EntMin+MAPLE with unlabeled dataset $D_u$.

\STATE \textbf{Initialization}: positive scalar weight ${\alpha}_{vat}$, ${\alpha}_{entmin}$ and ${\alpha}_{maple}$ for unsupervised loss $L_{vat}$, $L_{entmin}$ and $L_{maple}$, respectively; unlabeled batch size $b_u$, where $b_u \geq b_l$. 

\STATE \textbf{first repeat}
\INDSTATE t = 1 ... max iteration num:

\INDSTATE\hspace{\algorithmicindent} fetch mini-batch $d_l$ from $D_l$ and $d_u$ from $D_u$;

\INDSTATE\hspace{\algorithmicindent} compute loss $L = L_l + {\alpha}_{vat} L_{vat} + {\alpha}_{entmin} L_{entmin}$;

\INDSTATE\hspace{\algorithmicindent} update $\theta^t = \theta^{t-1} - \eta \bigtriangledown L$.

\STATE \textbf{until} stable accuracy and loss in the validation set.

\STATE \textbf{second repeat}
\INDSTATE t = 1 ... max iteration num:

\INDSTATE\hspace{\algorithmicindent} fetch mini-batch $d_l$ from $D_l$ and $d_u$ from $D_u$;

\INDSTATE\hspace{\algorithmicindent} compute loss $L = L_l + {\alpha}_{maple} L_{maple}$ on $d_l$ and $d_u$;

\INDSTATE\hspace{\algorithmicindent} update $\theta^t = \theta^{t-1} - \eta \bigtriangledown L$.

\STATE \textbf{until} stable accuracy and loss in the validation set.

\end{algorithmic}

\end{algorithm}

\subsection{Division of Semi-supervised Datasets }
Our experiments are conducted on three benchmark datasets: MSR-Action3D (MSR3D), NTU RGB+D 60 (NTU60), and NTU RGB+D 120 (NTU120). As shown in Table~\ref{supp_tab:table1}, we divide each training dataset into labeled training dataset $D_l$ and unlabeled training dataset $D_u$. As an illustration, we select 33 videos for each class (1980 videos in total) from NTU RGB+D 60 as the 5\% labeled training dataset. 

\subsection{Training Process of VAT+EntMin+MAPLE}

In this subsection, we describe the detailed training progress of ``VAT+EntMin+MAPLE''. As shown in Algorithm~\ref{supp_alg:algorithm1}, we first pre-train our model with labeled dataset $D_l$ for getting better initialization parameters. Then we adopt ${\alpha}_{vat} L_{vat} + {\alpha}_{entmin} L_{entmin}$ as the unsupervised loss with the unlabeled dataset $D_u$ until the model achieve stable accuracy (approximately 10 to 15 epochs for this step). At last, we use the $L_{maple}$ as our unsupervised optimization functions and continue training until the model almost converges.

\subsection{Results on NTU RGB+D 120 Dataset}

In this subsection, we additional evaluate our MAPLE by comparing it with leading semi-supervised methods on NTU RGB+D 120 dataset and show the results in Table~\ref{supp_tab:table2}. From the table, we can observe that previous semi-supervised methods (e.g. Pseudo Label, VAT and EntMin) are slightly effective for semi-supervised learning on this largest dataset of human action recognition. Our method outperforms the previous methods by about 1.0\% classification accuracy under the most setting.

\subsection{Reconstruction with Pseudo-Label}



\begin{figure*}[t]
\begin{minipage}{0.60\textwidth}

\subfigure[Exploding]{
\includegraphics[width=0.48\linewidth]{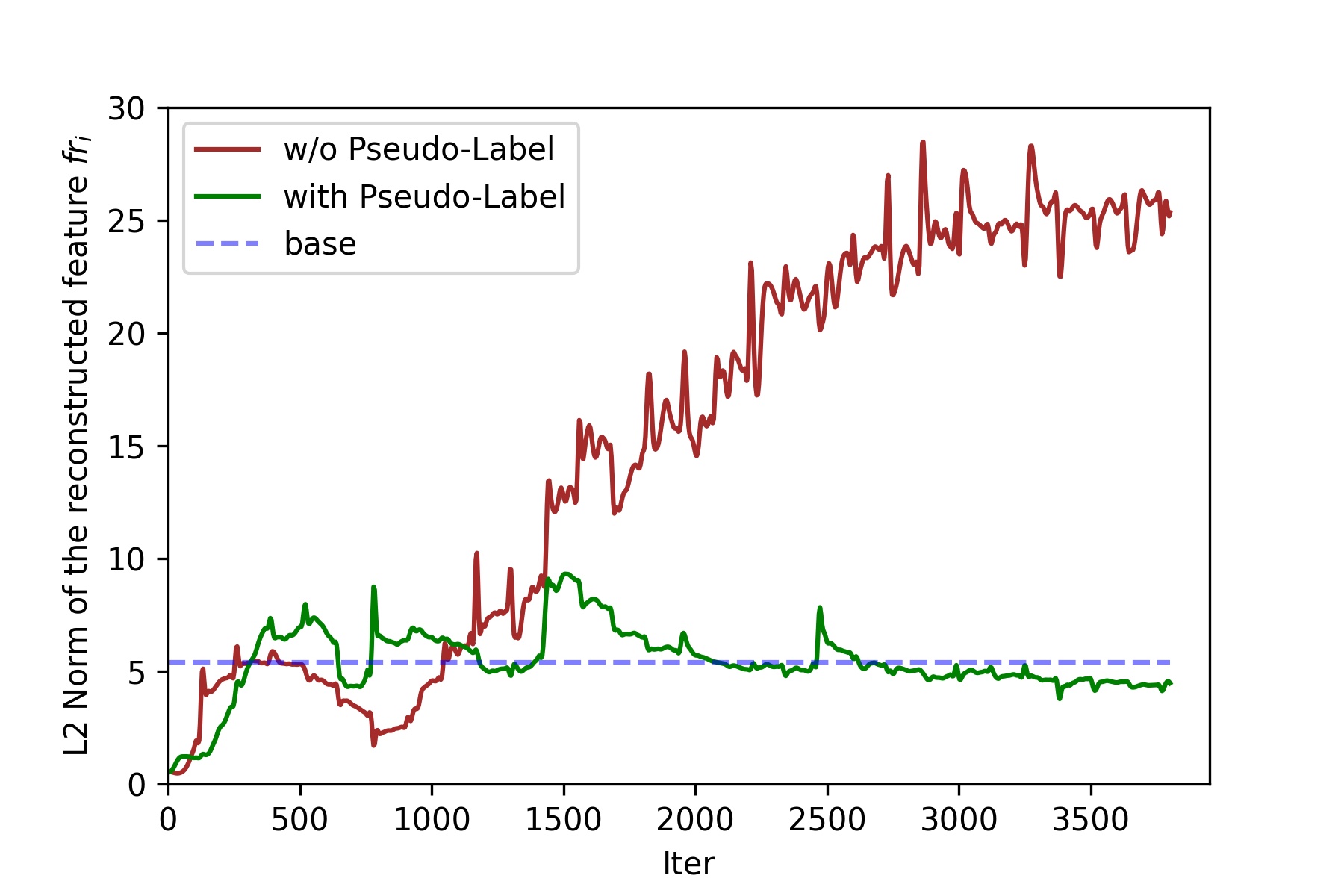}
}%
\subfigure[Vanishing]{
\includegraphics[width=0.48\linewidth]{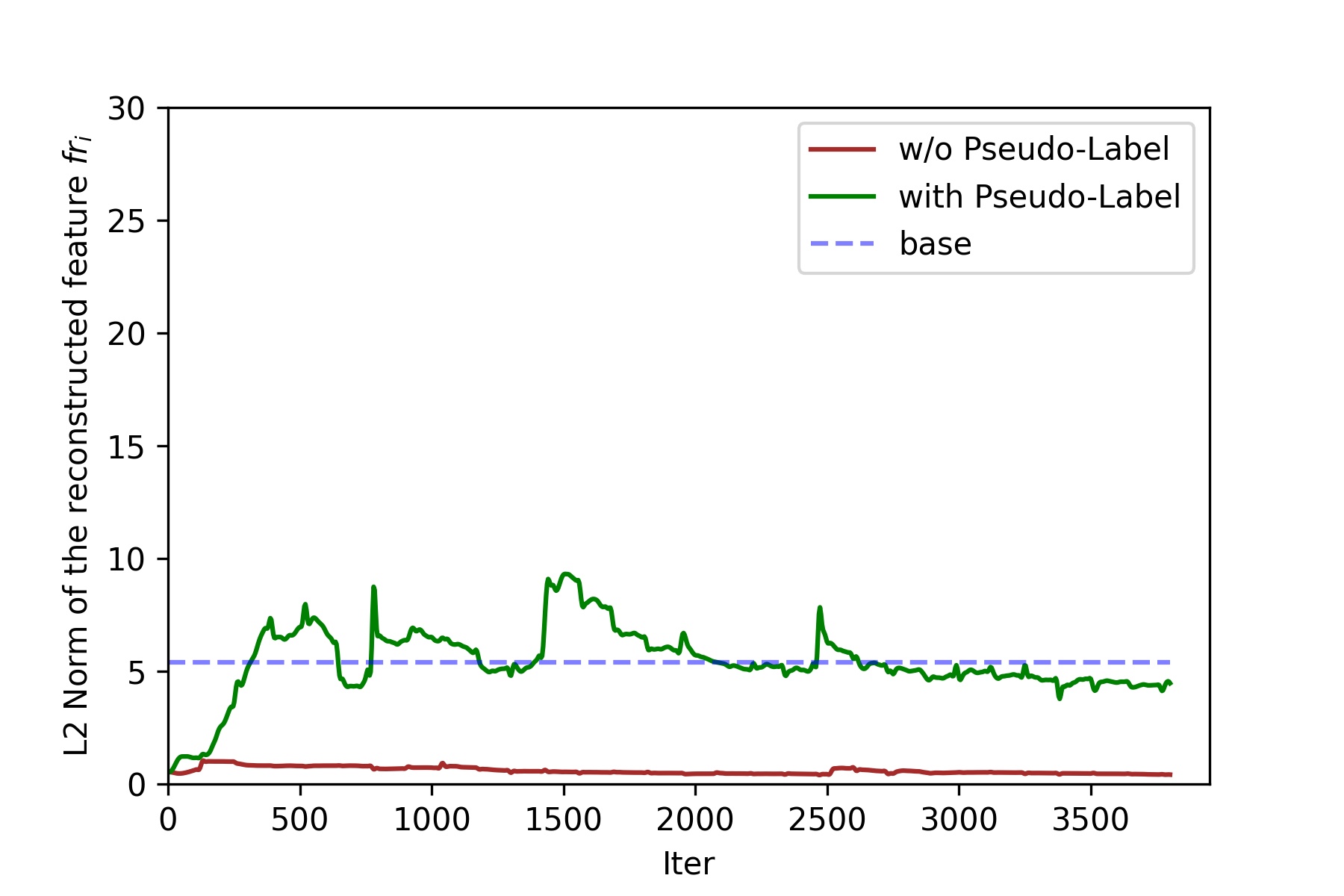}

}%
\caption{The $L2$ Norm of the reconstructed feature $r_i$ with and w/o pseudo-label in two common situations: (a) Exploding. (b) Vanishing. Note that the blue baseline denotes the average $L2$ Norm of the original feature $g_i$ during the supervised-only training process.}
\label{supp_fig:teaser1}
\end{minipage} \quad \begin{minipage}{0.33\textwidth}
  \includegraphics[width=0.99\linewidth]{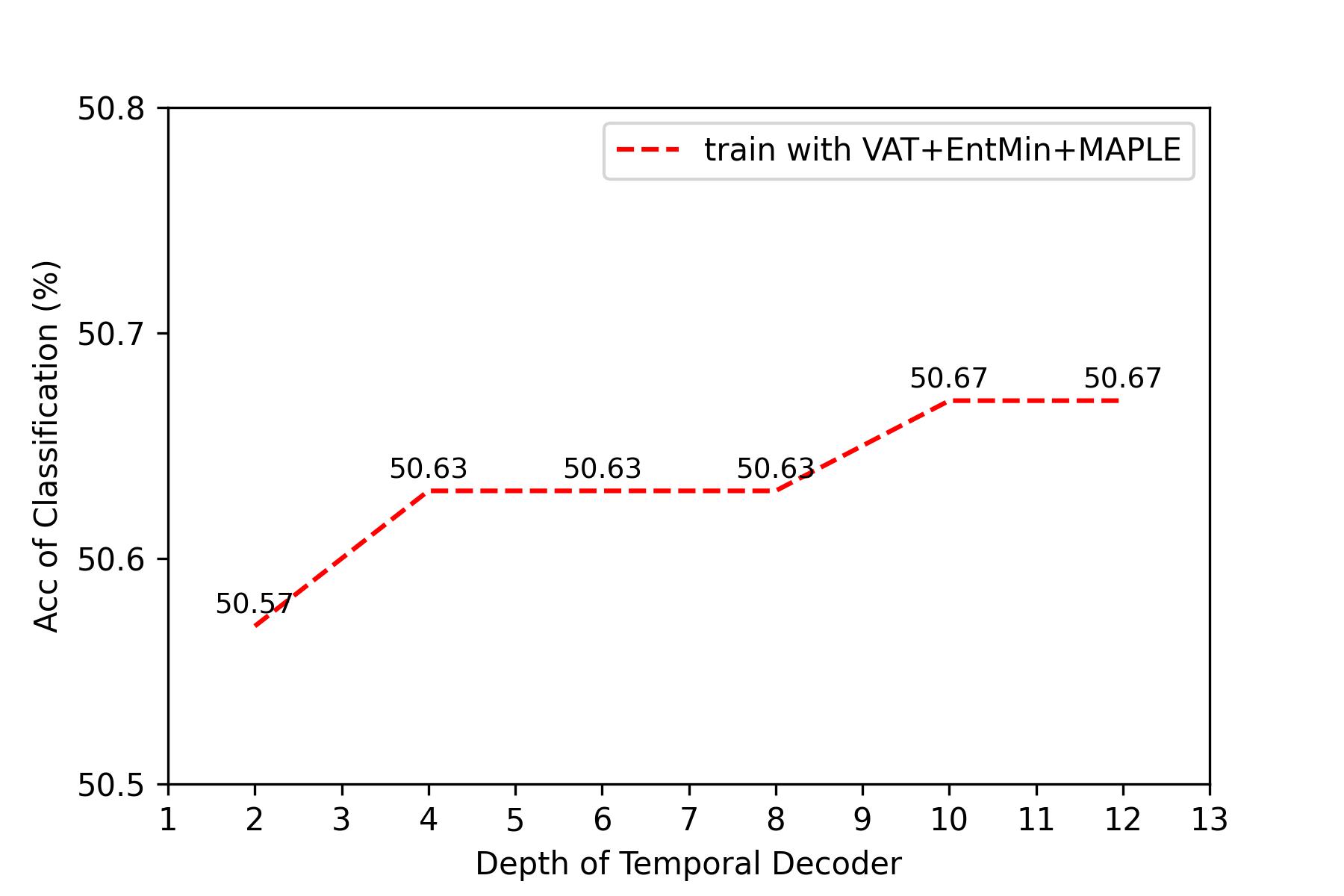}
  \caption{The accuracy of classification on NTU RGB+D 60 5\% labeled dataset with different depth of temporal decoder.}
  \label{fig:teaser2}
\end{minipage}
\end{figure*}

To demonstrate the importance of implicit reconstruction through pseudo-label, we show the $L2$ Norm of the reconstructed feature $r_i$ with and without pseudo-label on the 5\% labeled MSR-Action3D dataset in Fig.~\ref{supp_fig:teaser1} and compare the results of MAPLE under each setting in Table~\ref{supp_tab:table3}. The loss function without pseudo-label can be defined with Mean-Squared Error (MSE) loss as follows:
\begin{equation}
     L_u = L_{maple} =  \frac{1}{|D_u|} \sum_{x_i \in D_u} MSE( f(g_i|x_i) || f(r_i|x_i)) ,
\label{supp_equ:equ1}
\end{equation}
where $MSE$ is the function of MSE loss, $|D_u|$ is the size of the unlabeled dataset, $g_i$ is the original short-term global feature, and $r_i$ is the reconstructed short-term global feature. 
Note that the exploding and vanishing problem is not the same as gradient exploding and gradient vanishing. It indicates the difference in the feature size under each training strategy.

\begin{table*}
\centering
\begin{tabular}{l|ccccc} 
\hline
Action Accuracy (\%)    & high arm wave & horizontal arm wave & hammer       & hand catch   & forward punch   \\
supervised-only         & 6.67          & 86.67               & 0.00         & 33.33        & 91.67           \\
VAT+EntMin+MAPLE (Ours) & 93.33         & 86.67               & 0.00         & 26.67        & 21.43           \\ 
\hline
Action Accuracy (\%)    & high throw    & draw x              & draw tick    & draw circle  & hand clap       \\
supervised-only         & 21.43         & 76.92               & 100.00       & 6.67         & 84.62           \\
VAT+EntMin+MAPLE (Ours) & 28.57         & 42.86               & 100.00       & 73.33        & 100.00          \\ 
\hline
Action Accuracy (\%)    & two hand wave & side-boxing         & bend         & forward kick & side kick       \\
supervised-only         & 100.00        & 35.71               & 20.00        & 100.00       & 100.00          \\
VAT+EntMin+MAPLE (Ours) & 100.00        & 100.00              & 80.00        & 100.00       & 100.00          \\ 
\hline
Action Accuracy (\%)    & jogging       & tennis swing        & tennis serve & golf swing   & pick up  throw  \\
supervised-only         & 100.00        & 86.67               & 93.33        & 46.67        & 66.67           \\
VAT+EntMin+MAPLE (Ours) & 100.00        & 93.33               & 100.00       & 73.33        & 93.33           \\
\hline
\end{tabular}
\caption{More details about the improvement of per class accuracy on the 7.5\% labeled MSR-Action3D dataset.}
\label{supp_tab:table4}
\end{table*}

From the figure and table, we can observe that training without pseudo-label lead to explosion or vanishment on the $L2$ Norm of the reconstructed features, which not only results in a significant decrease in the final classification performance but also make it hard to stably converge during the training process. 
The main reason for this occurrence can be obtained from our loss function~\ref{supp_equ:equ1} and the encoder-decoder structure of our MAPLE. 
The explosion situation (Fig.~\ref{supp_fig:teaser1} (a)) happens when both the detached $g_i$ (without backprop) and $r_i$ are learnable features generated by our MAPLE and they share the same spatial extractor. When $r_i$ tries to increase closer to the  $g_i$, it often leads to an increase in the weights of the spatial extractor, which in turn leads to the increase of $g_i$. After hundreds or thousands of iterations, the $L2$ Norm of $g_i$ and $r_i$ become larger and larger, even tending to infinity.
The vanishment situation (Fig.~\ref{supp_fig:teaser1} (b)) happens when both the original $g_i$ (with backprop) and $r_i$ are learnable features generated by our MAPLE. The model tends to simply reduce the $L2$ Norm of $g_i$ and $r_i$ to achieve the reduction of loss function~\ref{supp_equ:equ1}. After thousands of iterations, the $L2$ Norm of $g_i$ and $r_i$ become smaller and smaller, even tending to zero.

\subsection{Depth of Temporal Decoder}


We investigate the effect of decoder depth on the final classification performance and the result on NTU RGB+D 60 5\% labeled dataset is shown in Fig.~\ref{fig:teaser2}. From the figure we can know that the depth of the temporal decoder does not have a large impact on the performance of the final classification, the main reason is that the role of the decoder is only used to reconstruct the complete action sequence from the latent space, and has less relevance to the classification. Therefore, a decoder of shallow depth is sufficient to reconstruct the complete sequence.

\subsection{Details about per class accuracy}
To find out what type of actions are easily classified and which ones are tough, we show more details about the improvement of per class accuracy after training with our MAPLE in Table~\ref{supp_tab:table4} on the 7.5\% labeled MSR-Action3D dataset. From an overall perspective, it seems that simple repetitive actions, such as high arm wave (+86.67\% accuracy), can greatly benefit from our MAPLE, while some complex actions, such as draw x (-34.06\% accuracy), are hardly benefit from our MAPLE.



\end{document}